\documentclass[10pt,twocolumn,letterpaper]{article}

\usepackage{cvpr}
\usepackage{times}
\usepackage{epsfig}
\usepackage{graphicx}
\usepackage{amsmath}
\usepackage{amssymb}
\usepackage{enumitem}
\usepackage{color}
\usepackage{verbatim}
\usepackage{bm}
\usepackage{array}
\usepackage{caption}
\usepackage{subfig}

\newcommand{\beq}{\begin{equation}}
\newcommand{\eeq}{\end{equation}}

\newif\ifteaser
\teasertrue

\usepackage[pagebackref=true,breaklinks=true,letterpaper=true,colorlinks,bookmarks=false]{hyperref}

\def\cvprPaperID{---} 

\cvprfinalcopy
\ifcvprfinal\fi
\begin{document}

\title{{L}ocally {N}on-rigid {R}egistration for {M}obile {HDR} {P}hotography}

\author{
Orazio Gallo$^{1}$\hspace{1.5em}
Alejandro Troccoli$^{1}$\hspace{1.5em}
Jun Hu$^{1,2}$\hspace{1.5em}
Kari Pulli$^{1,3}$\hspace{1.5em}
Jan Kautz$^{1}$\hspace{1.5em}\\
\small{$^{1}$NVIDIA\hspace{1.5em} $^{2}$Duke University \hspace{1.5em} $^{3}$Light}}


\ifteaser
\twocolumn[{%
\renewcommand\twocolumn[1][]{#1}%
\maketitle
\begin{center}
    \centering
    \includegraphics[width=0.99\textwidth]{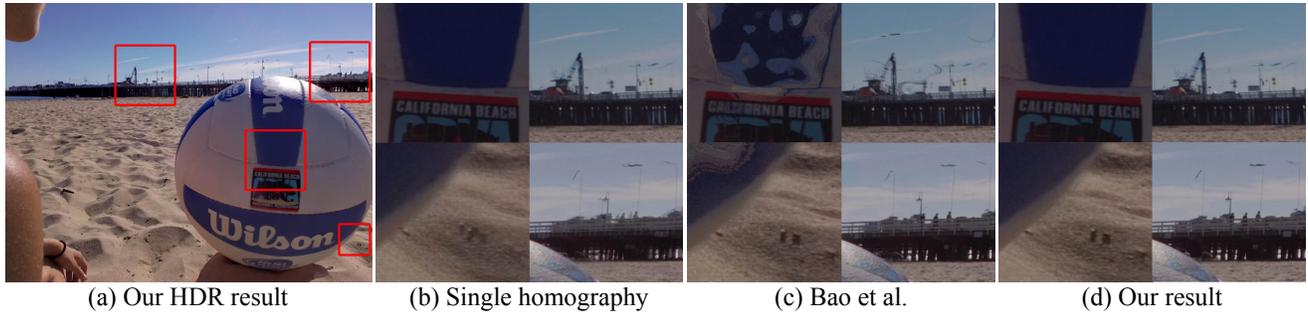}
    \captionof{figure}{\small{When capturing HDR stacks without a tripod, parallax and non-rigid scene changes are the main sources of artifacts. The picture in (a) is an HDR image generated by our algorithm from a stack of two pictures taken with a hand-held camera (notice that the volleyball is hand-held as well). A common and efficient method to register the images is to use a single homography, but parallax will still cause ghosting artifacts, see (b). One can then resort to non-rigid registration methods; here we use the fastest method of which we are aware, but artifacts due to erroneous registration are still visible (c). Our method is several times faster and, for scenes with parallax and small non-rigid displacements, produces better results (d).}}\label{fig:teaser}
\end{center}%
}]
\else
\maketitle
\fi

\begin{abstract}
Image registration for stack-based HDR photography is challenging. If not properly accounted for, camera motion and scene changes result in artifacts in the composite image. Unfortunately, existing methods to address this problem are either accurate, but too slow for mobile devices, or fast, but prone to failing. We propose a method that fills this void: our approach is extremely fast---under $700$ms on a commercial tablet for a pair of 5MP images---and prevents the artifacts that arise from insufficient registration quality.
\end{abstract}

\section{Introduction}

High-Dynamic-Range (HDR) imaging has become an essential feature for camera phones and point-and-shoot cameras---even some DSLR cameras now offer it as a shooting mode. To date, the most popular strategy for capturing HDR images is to take multiple pictures of the same scene with different exposure times, which is usually referred to as \emph{stack-based} HDR. Over the last decade, the research community also proposed several hardware solutions to sidestep the need for multiple images, but the trade-off between cost and picture quality still makes stack-based strategies more appealing to camera manufacturers.

Combining multiple low-dynamic-range (LDR) images into a single HDR irradiance map is relatively straightforward, provided that each pixel samples the exact same irradiance in each picture of the stack. In practice, however, any viable strategy to merge LDR images needs to cope with both camera motion and scene changes. Indeed there is a rich literature on the subject, with different methods offering a different compromise between computational complexity and reconstruction accuracy.

\ifteaser\else
\begin{figure}
\centering
\includegraphics[width=0.85\columnwidth]{Figures/teaser/teaser.jpg}\\
\includegraphics[width=0.85\columnwidth]{Figures/teaser/all_details.pdf}
\caption{\protect\input{teaserCaption}}\label{fig:teaser}
\end{figure}
\fi

On one end of the spectrum there are light-weight methods, generally well-suited to run on mobile devices. These methods address the problem of camera motion by estimating a global transformation in a robust fashion~\cite{Ward2003,Tzimiropoulos2010}. After image alignment, scene changes can be addressed with some flavor of outlier rejection, often called deghosting. This can be achieved by picking one image of the stack to act as a reference and only merging consistent pixels from the other images~\cite{Gallo2009,Raman2011}. Alternatively, for sufficiently large stacks, one can merge only the irradiance values most often seen for a given pixel~\cite{Zhang2012,Oh2014}. The price for the computational efficiency of rigid-registration methods is their severe limitation in terms of accuracy: even the most general global transformation, \ie, a homography, cannot correct for parallax, which occurs for non-planar scenes every time the camera undergoes even a small amount of translation.

On the other end of the spectrum lie methods that allow for a completely non-rigid transformation between the images in the stack~\cite{Sen2012,Hu2013}. Rather than separating camera motion and scene changes, these algorithms attempt to ``move'' any given pixel in one shot of the stack to its corresponding location in the reference image. These methods have shown impressive results, essentially with any amount and type of motion in the scene, but generally require minutes on desktop computers: they are simply impractical for deployment on mobile devices.

We fill the gap between these two extreme points in the space of registration accuracy versus computational complexity.  Our work builds on the observation that most modern devices, such as the NVIDIA SHIELD Tablet or the Google Nexus 6 phone, are capable of streaming full-resolution YUV frames at $30$fps. Given an exposure time $t$, the delay between consecutive shots is then $(33-t)\mbox{ms}<33\mbox{ms}$\footnote{If $t > 33\mbox{ms}$, the limiting factor will likely be blur, and the delay between shots will still be $(33-\mathbin{\mbox{mod}}(33,t))\mbox{ms} <33\mbox{ms}$.}, which prevents large displacements of moving objects. Parallax, however, remains an issue even for small camera translations. Figure~\ref{fig:teaser}(b) shows the extent of these artifacts. (Note that, for better visualization, all the insets in Figure~\ref{fig:teaser} were generated with a simple blending.)

Our method comprises a strategy to find sparse correspondences that is particularly well-suited for HDR stacks, where large parts of some images are extremely dark. After detecting spurious matches, it corrects for parallax by propagating the displacement computed at the discrete locations in an edge-aware fashion. The proposed algorithm can also correct for small non-rigid motions, but may fail for cases where the subject simply cannot be expected to cooperate, as is the case in sport photography. To address such cases, we couple our locally non-rigid registration algorithm with a modified version of the exposure fusion algorithm~\cite{Mertens2007}.

Our method runs in $677$ms on a stack of two 5MP images \emph{on a mobile device}, which is several orders of magnitude faster than any non-rigid registration method of which we are aware. On a desktop machine, our method can register the same stack in $150$ms, corresponding to a speedup of roughly $11\times$ over the fastest optical flow methods published to date (see Section~\ref{sec:results}).

\section{Method}

Several recently published methods successfully tackled the task of non-rigid registration for large displacements by using approximate nearest neighbor fields (NNFs)~\cite{Hu2013,Sen2012}. While the quality of the results they produce is impressive, even in the case of very large displacements, their computational cost is prohibitive, in particular for mobile devices. Moreover, given the frame rate at which bursts of images can be acquired, the tolerance to large displacements that those methods offer is most often unnecessary.

The problem of large displacements is further attenuated by the dynamic range of modern sensors, which allows to capture most scenes with only two shots; leveraging on this observation, we focus on two-image exposure stacks, although the extension to more images only requires to run the algorithm $n-1$ times for a stack of $n$ images, as shown in Figure~\ref{fig:fullNonRigidComparison}, where we register a stack of three images. Rather than computing an expensive NNF, which, for the vast majority of stacks, would mostly consist of small and relatively uniform displacements, we find sparse correspondences between the two images.
While extremely fast, the matcher we designed for this purpose produces accurate matches, even in extremely dark regions---a particularly important feature for HDR stacks. To solve the parallax problem, rather than registering the images with a single homography, we propose to propagate the sparse flow from the matches computed  in the previous stage in an edge-aware fashion.
To merge the images we modified exposure fusion~\cite{Mertens2007} to compensate for potential errors in the computation of the flow.
We implemented the full pipeline---stack capture, image registration, and image fusion---on an NVIDIA SHIELD Tablet.

In the remainder of this section we describe in detail the different components of our algorithm.

\subsection{Stack capture and reference selection}

Metering for HDR, \ie, the selection of the exposure times and number of pictures required to sample the irradiance distribution for a particular scene, has been an active area of research~\cite{Gallo2012,Granados2010,Hasinoff2010}. We observe that the dynamic range of modern sensors allows to capture most real-world scenes with as little as two exposures, and devise a simple strategy that works well in our experiments: we use the Expose-To-The-Right (ETTR) paradigm~\cite{Hasinoff2010} for the first image in the stack, and select the second exposure time to be $2$, $3$, or $4$ stops brighter, based on the number of under-exposed pixels (the more under-exposed pixels, the longer the second exposure). Limiting the number of candidate exposures to three allows for a faster metering; moreover, the advantage of a higher granularity is difficult to appreciate by visually inspecting the HDR result.

Then, rather than picking the reference image for our registration algorithm to be the one with the least saturated and underexposed pixels~\cite{Gallo2009}, we always use the shortest (darkest) exposure as the reference; this is because, while the noise in the dark regions of a scene makes it difficult to find reliable matches, saturation makes it impossible. In the rest of the paper we will refer to the two images in the stack as \emph{reference} and \emph{source}, indicating our final goal to warp the source to the reference.

\subsection{A fast, robust matcher}\label{sec:matcher}

\begin{figure}[!t]
\centering
\includegraphics[width=.7\columnwidth]{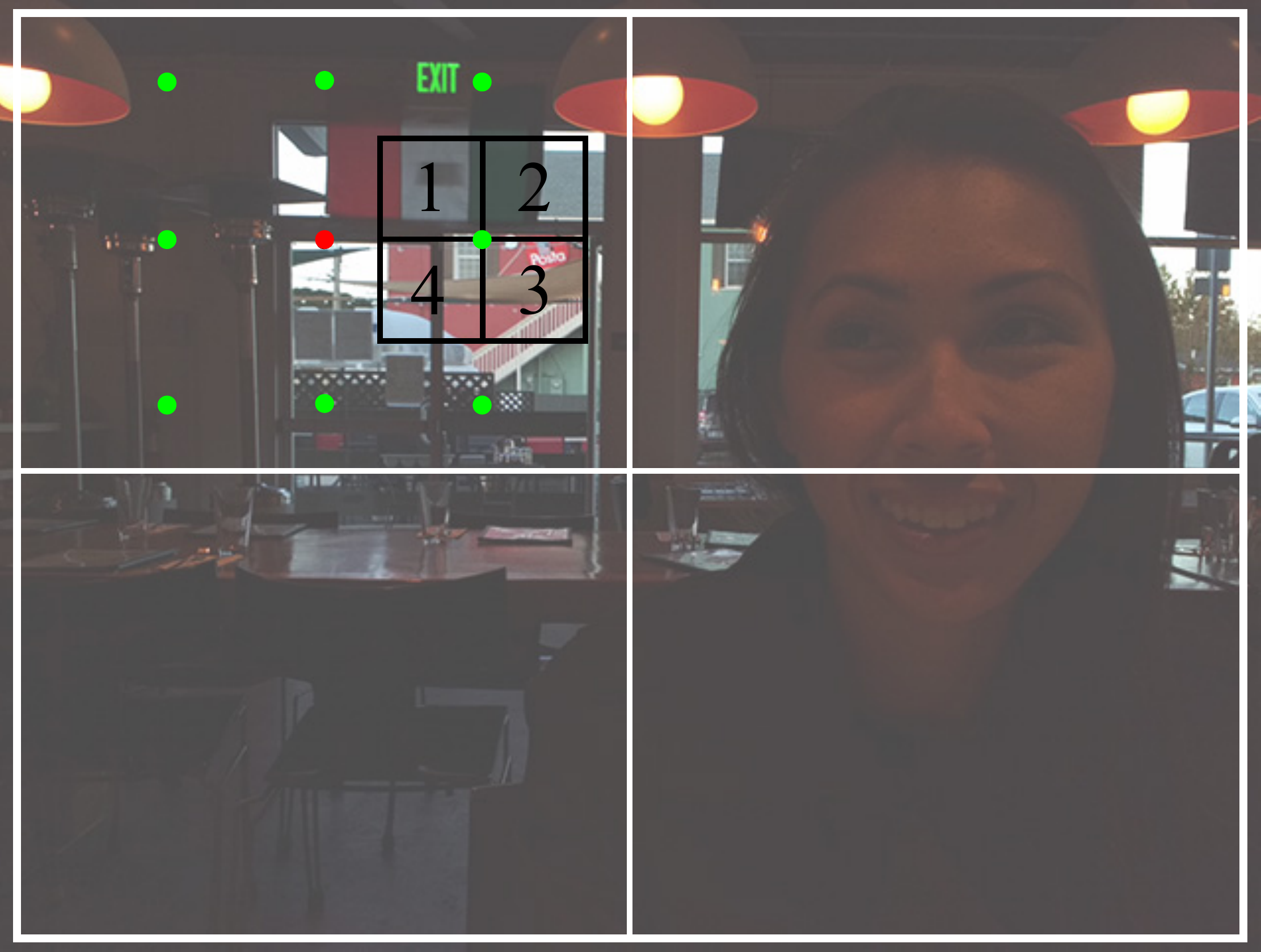}
\caption{After dividing the reference image in tiles, our matcher looks at the center of each tile (red dot) and a set of predefined locations around it (green dots). It then computes a measure of \emph{cornerness} based on the average luminance in the four quadrants around each candidate corner.}\label{fig:matcher}
\end{figure}

In order to produce fast and reliable correspondences, even in the presence of the noise the potentially short exposure time induces, we propose a novel matcher.
To this end, we efficiently find distinct features, \ie, corners, in the reference image, which we then match in the second image with a patch-based search. 
First, we define a simple measure of \emph{cornerness}:
\beq\label{eq:cornerness}
C(p) = \sum_{j=1}^4  |\mu_{\text{mod}(j+1,4)} - \mu_j|
\eeq
where $p$ is the pixel location, and $\mu_j$ is the average luminance value in the $j^{\text{th}}$ quadrant around $p$, marked in black in Figure~\ref{fig:matcher}.
Equation~\ref{eq:cornerness} simply measures the change in the average luminance in the four quadrants around $p$. However, for $p$ to be a good corner candidate, we also require that the minimum difference of average luminance between any two contiguous quadrants be large:
\beq\label{eq:minimumActivity}
 \min\limits_{j \in \{1, \ldots, 4\}}  |\mu_{\text{mod}(j+1,4)} - \mu_j| > T
\eeq
where $T$ is a given threshold. Essentially, Equation~\ref{eq:minimumActivity} prevents situations in which a point on an edge is promoted to a corner.
Note that, regardless of the number of pixels in each quadrants, $\mu_j$ can be computed very efficiently using integral images.

To encourage a uniform distribution of corners over the image, we first divide the reference image in tiles. Then, we look at a predefined set of locations (and not at all pixels for computational efficiency) around the center of each tile (Figure~\ref{fig:matcher}), and retain the one with the highest cornerness value.
Note that Equation~\ref{eq:minimumActivity} also prevents finding corners in flat regions, therefore tiles that are completely flat may not be assigned a corner.

Although the proposed algorithm to search for corners is extremely efficient, we do not run it on both reference and source images, as it only serves as a feature detector.
Hence, we evaluate corners in the reference image only, and then search for matches within a predefined search radius around the corresponding position in the source image using the sum of squared differences metric (SSD).

Finally, to minimize the computational cost while allowing for a large search radius, we use a pyramidal approach. At level $\ell$ of the pyramid, we find a set of corners $\boldsymbol{x}^\ell_{\text{ref}}$ in the reference. For each corner $\boldsymbol{x}^\ell_{\text{ref}}$ we search for matches in the source image within a search radius around
\beq\label{eq:searchInit}
\boldsymbol{x}^\ell_{\text{src}} = H^{\ell+1} \boldsymbol{x}^\ell_{\text{ref}},
\eeq
where $H^{\ell+1}$ is a single homography computed at the previous layer of the pyramid using its corners and matches. Equation~\ref{eq:searchInit} holds if both $\boldsymbol{x}_{\text{src}}$ and $\boldsymbol{x}_{\text{ref}}$ are represented in a normalized coordinate system such that $x \in [-1,1]$ and $y \in [-h/w,h/w]$, where $\boldsymbol{x}= (x,y)$, and $w$ and $h$ are width and height of the image. We also move the origin to the center of the image. This homography
only serves as a way to initialize the search locations in the source image (and in turn reduces the required search radius compared to a random initalization).
Note that $\boldsymbol{x}^\ell_{\text{ref}}$ are computed directly on layer $\ell$, and not upsampled from layer $\ell+1$, as relevant features at layer $\ell$ may have not been present at layer $\ell+1$ and may not be visible in layer $\ell-1$.

\subsection{Weeding out spurious  matches}\label{sec:cleaning}

The matcher we describe in Section~\ref{sec:matcher} generally produces robust matches even in very low light conditions (see Section~\ref{sec:results} for a more detailed evaluation). However, to detect and remove potential spurious matches, we run an additional filtering stage.
The key idea is that we require matches to be locally consistent with a homography; those that are not, are likely to be incorrect, because we expect small displacements. The consistent matches can be determined by means of a robust method, such as RANSAC~\cite{RANSAC}. A straightforward application of RANSAC, however, is too expensive. Instead, we developed a fast filtering strategy to weed out spurious matches that are not consistent with a homography.

The goal of our novel filter is to efficiently obtain the set $M$ of reliable matches. We consider a match to be reliable if there exists a large number of matches that are mapped from the source image to the reference image by the same local homography $H$; in other words, we aim at finding all the matches that induce any homography supported by a large set of inliers.

The filter, inspired by RANSAC, works iteratively. Specifically, at the $i^{\text{th}}$ iteration, we randomly sample $4$ points from our set of matches, fit a homography $H^i$, and find the subset of inliers $I^i$ with respect to $H^i$. Then, rather than saving the homography supported by the largest set of inliers, we simply update $M$ using the following rule:
\beq\label{eq:update}
M^i = \begin{cases} M^{i-1} \cup I^i &\mbox{if } |I^i| >\delta \\ 
M^{i-1} & \mbox{otherwise}\end{cases}~~~, 
\eeq
where $|\cdot|$ indicates the cardinality of a set, and $\delta$ is a threshold.
To understand the idea behind Equation~\ref{eq:update}, consider a toy scene where most of the corners are distributed on two static planar surfaces at different distances from the camera, with a few other moving non rigidly. We would like our algorithm to weed out the non-rigid corners, as well as the corners on the two surfaces that are incorrectly matched. At each iteration, one of two things can happen. First, the sampling may include corners from both planes or those moving non-rigidly; the resulting homography will have a small number of inliers, and the set of reliable matches $M$ will not be modified. Second, all of the points sampled at the current iteration belong to one of the two planes. In this case the resulting homography will explain the motion of all the corners that are on the same plane and that move rigidly; the set of reliable matches $M$ will be updated to include these inliers. Note that we do not remove the inliers of the $i^{\text{th}}$ iteration from the original set of corners.

We further speed up the process by running $n$ instances of our filter2 on separate threads, with each instance running only $1/n$ iterations; because we take the union of the acceptable inliers from previous iterations, running our filter $n$ times for $1/n$ iterations is exactly equivalent to a single run on $n$ iterations. After each run $k$ terminates, we simply merge the resulting sets $M_k$.

\subsection{Sparse-to-dense flow}\label{sec:flowPropagation}
So far, we have described a method to efficiently find a set of robust matches between the images, constituting sparse flow. To be able to warp the source image to the reference, however, we need to compute the displacement at every pixel. A simple interpolation of the sparse flow would produce artifacts similar to those caused by using a single homography: depth discontinuities and boundaries of moving objects would not be aligned accurately.

Instead, we would like to interpolate the sparse flow in an edge-aware fashion. The problem is similar to that of image colorization~\cite{Levin2004}, where colors are propagated from a handful of sparse pixels that have been assigned a color manually.
In our case, we propagate the flow components $(u,v)$ computed at discrete locations.

For this purpose, we employ an efficient CUDA implementation of the algorithm proposed by Gastal and Oliveira~\cite{Gastal2011}, and use it to cross-bilateral filter the flow. 
Similarly to how they propose to propagate colors, we first create two maps $P_u$ and $P_v$
\beq\label{eq:flowMaps}
P_f(p) = \begin{cases} f(p) &\mbox{if $p$ is a corner} \\ 
0 & \mbox{elsewhere}\end{cases}~~~, 
\eeq
where $f = \{u,v\}$.
We then use the reference image to cross-bilateral filter the maps.
However, while this propagates the flow in an edge-aware fashion generating the two maps $\widetilde{P}_f$, it will affect the value of the flow at the location of the corners, which should not change.
Therefore, we use a normalization map
\beq\label{eq:normMaps}
N(p) = \begin{cases} 1 &\mbox{if $p$ is a corner} \\ 
0 & \mbox{elsewhere}\end{cases}~~~.
\eeq
The final flow $F$ can then be computed as $F_f = \widetilde{P}_f/\widetilde{N}$, where $\widetilde{N}$ is the cross-bilateral filtered version of $N$.

\subsection{Error-tolerant image fusion}\label{sec:ssim}
If the number of matches between reference and source images is low, or if a particular area of the scene is textureless, the quality of the flow propagation described in Section~\ref{sec:flowPropagation} can deteriorate because the accuracy of the flow is affected by the spatial distance over which it needs to propagate.
To detect and compensate for errors that may arise in such cases, we propose a simple modification of the exposure fusion algorithm proposed by Mertens~\etal~\cite{Mertens2007}. In addition to weights for contrast, color saturation, and well-exposedness, we add a fourth weight that reflects the quality of the registration. Specifically, we choose to use the structural similarity index (SSIM)~\cite{Wang2004}. Note that computing the SSIM map only requires to perform five convolutions with Gaussian kernels and a few other parallelizable operations such as pixel-wise image multiplication and sum; this makes a GPU implementation of SSIM extremely efficient, see Section~\ref{sec:results} for an analysis of its runtime.

\begin{figure}%
\centering
\subfloat[Blended stack]{\includegraphics[width=0.315\columnwidth]{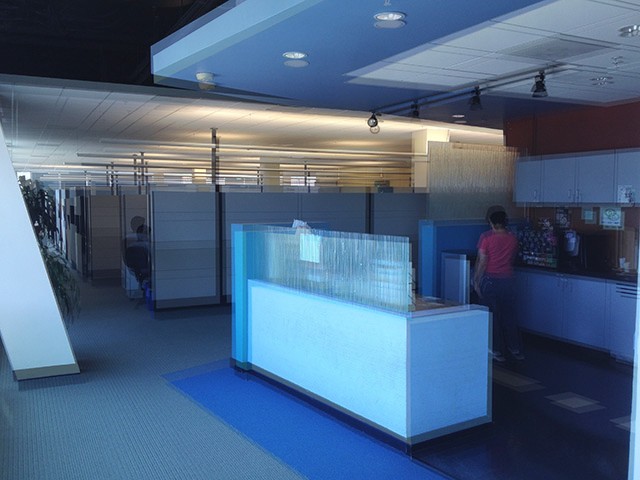}}\enskip
\subfloat[Source warped]{\includegraphics[width=0.315\columnwidth]{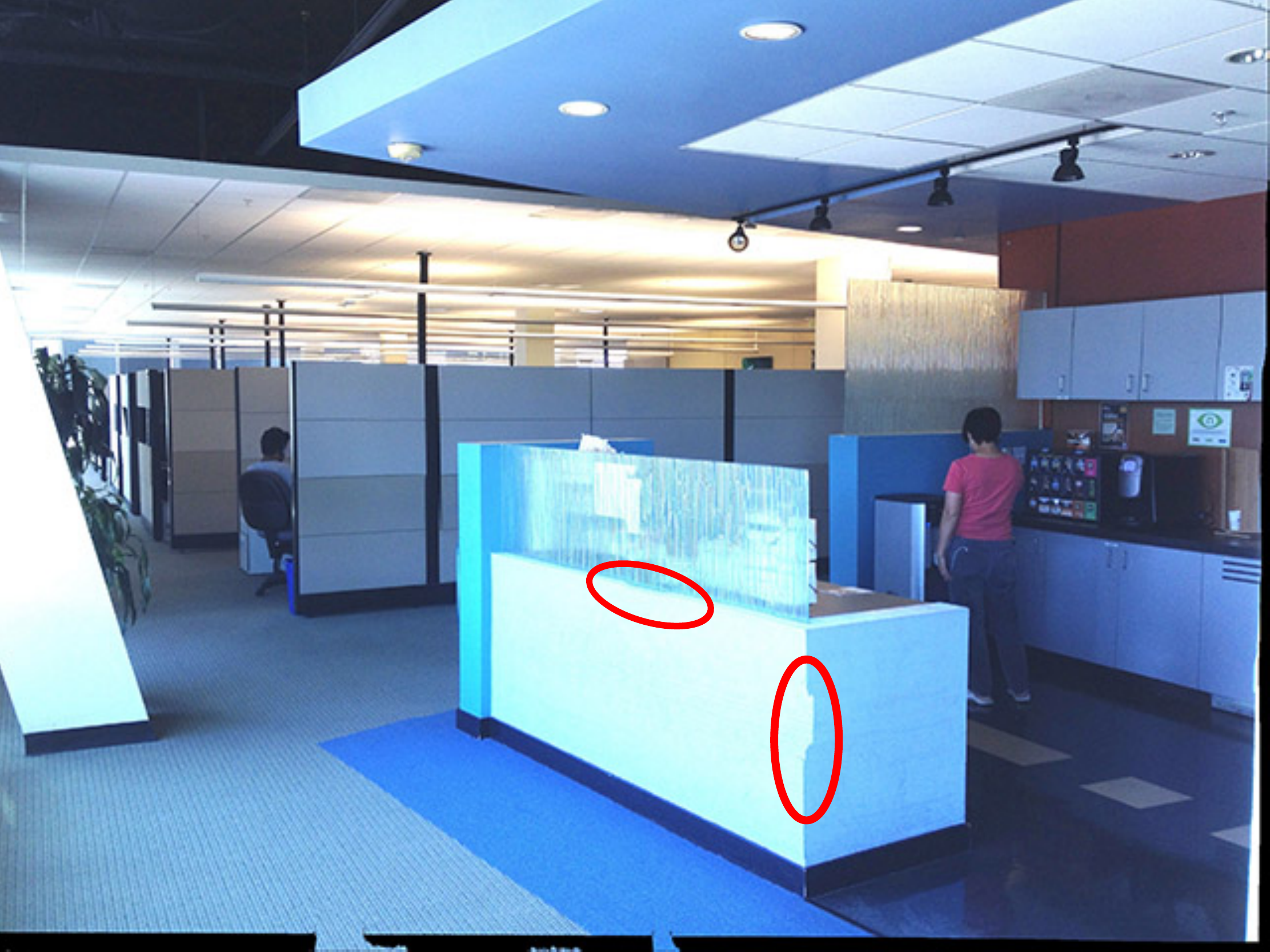}}\enskip
\subfloat[Final HDR]{\includegraphics[width=0.315\columnwidth]{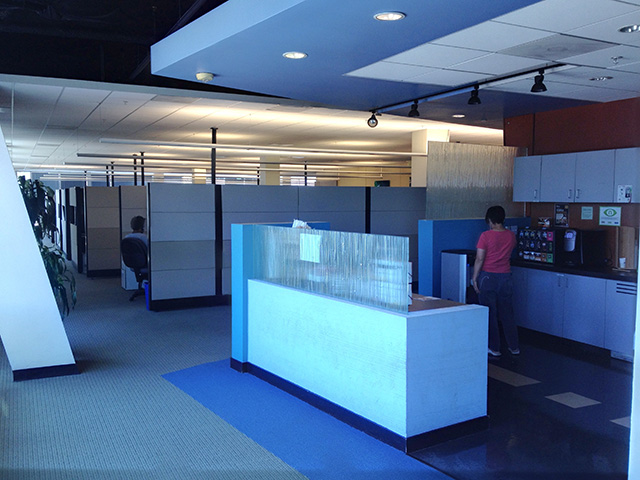}}
\caption{A failure case of the flow propagation stage. The input images are first blended to show the original displacement (a). The warped source produced by our algorithm presents a few artifacts, a couple of which are marked. However, our error-tolerant exposure fusion can detect and correct for those errors.}\label{fig:fail}
\end{figure}

Figure~\ref{fig:fail} shows an example of failure of the edge-aware propagation stage, and how our error-tolerant fusion can detect and compensate for it.

\subsection{Implementation details}
For the matcher, we create up to $5$ pyramid layers (we stop early if the coarsest level falls below $100$ pixels in either height or width). The patch comparison is computed on $21\times21$ patches, and the maximum search radius at each level is $10$. We evaluate the cornerness within a patch on a regular grid of points spaced by $1/16^{\text{th}}$ of the tile size.
We implemented the method described here---from capture to generation of the HDR image, with a mixture of C++ and CUDA code. Specifically, the matcher and the weeding stage run on the CPU, with the weeding stage being multi-threaded. The remaining parts are heavily CUDA-based. Finally, for the sparse-to-dense stage, it is important to use a large spatial standard deviation to make sure that the flow can be propagated to regions poor in number of correspondences; we use $\sigma_s = 400$ .

\section{Evaluation and Results}\label{sec:results}

In this section we evaluate the performance of our algorithm, both in terms of quality of the result and execution time, by means of comparisons with state-of-the-art methods. Note that we perform histogram equalization on all the input images to attenuate brightness differences.

\begin{figure*}%
\centering
\subfloat[Reference]{\includegraphics[width=0.119\textwidth]{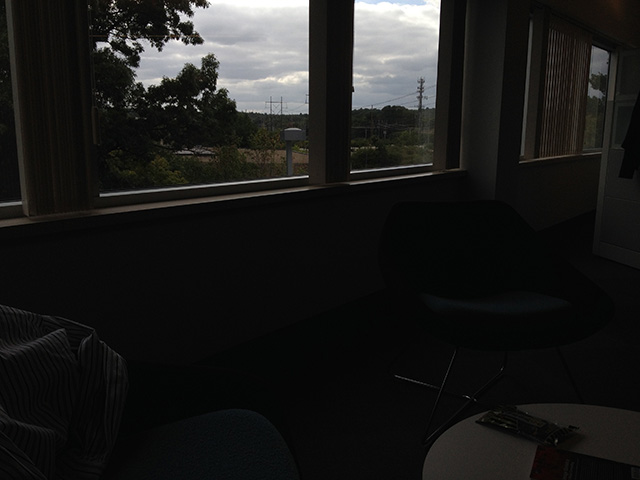}}\enskip
\subfloat[Source]{\includegraphics[width=0.119\textwidth]{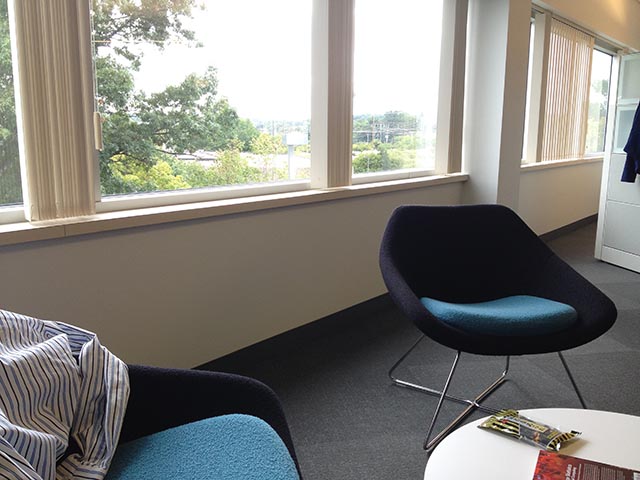}}\enskip
\subfloat[SIFT]{\label{fig:siftMatches1}\includegraphics[width=0.238\textwidth]{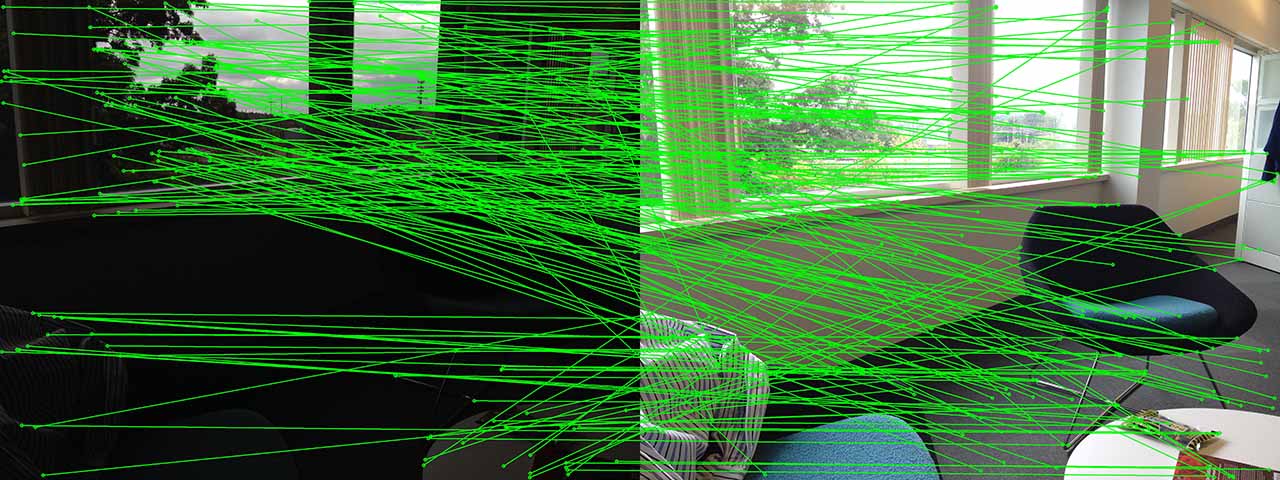}}\enskip
\subfloat[SIFT clean]{\label{fig:siftMatchesClean1}\includegraphics[width=0.238\textwidth]{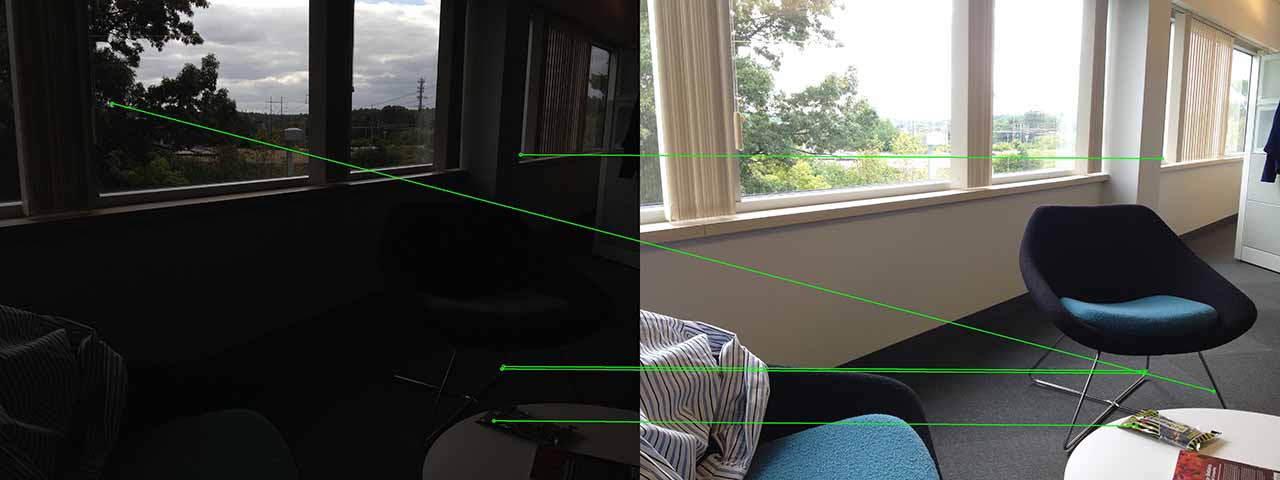}}\enskip
\subfloat[Ours]{\includegraphics[width=0.238\textwidth]{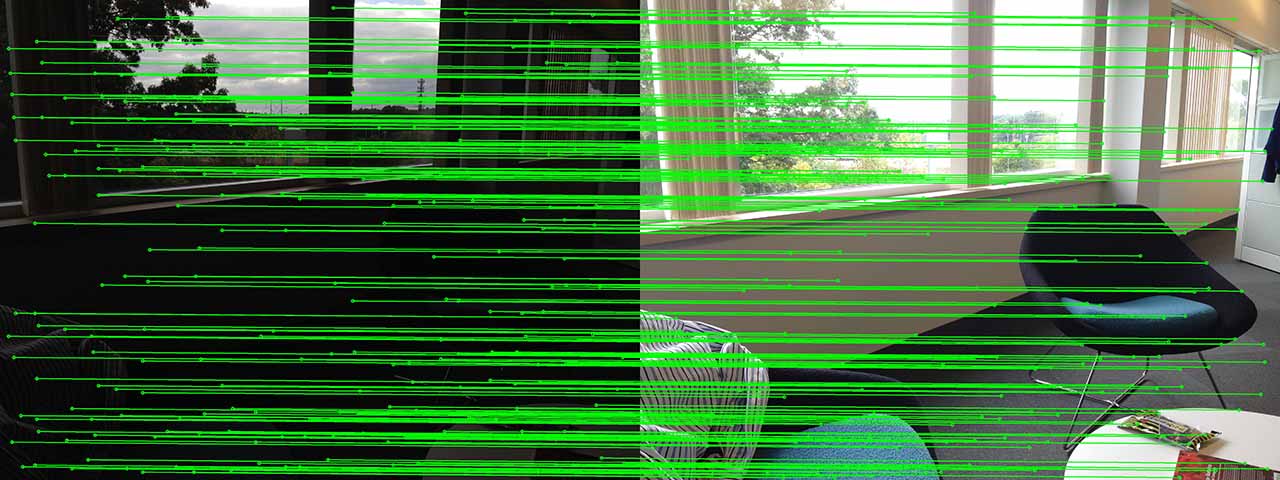}}\\
\subfloat[Reference]{\includegraphics[width=0.119\textwidth]{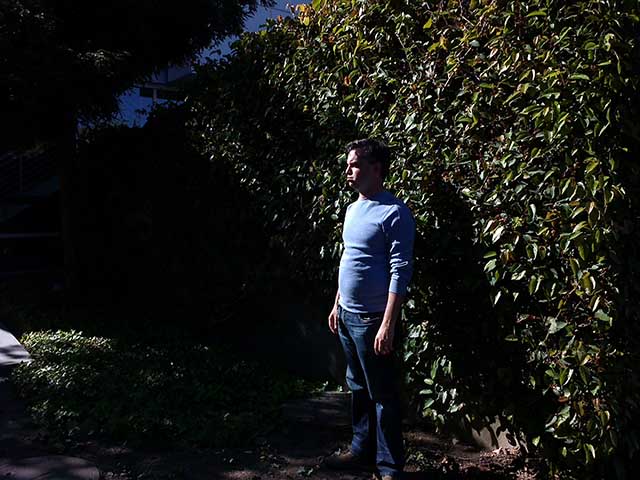}}\enskip
\subfloat[Source]{\includegraphics[width=0.119\textwidth]{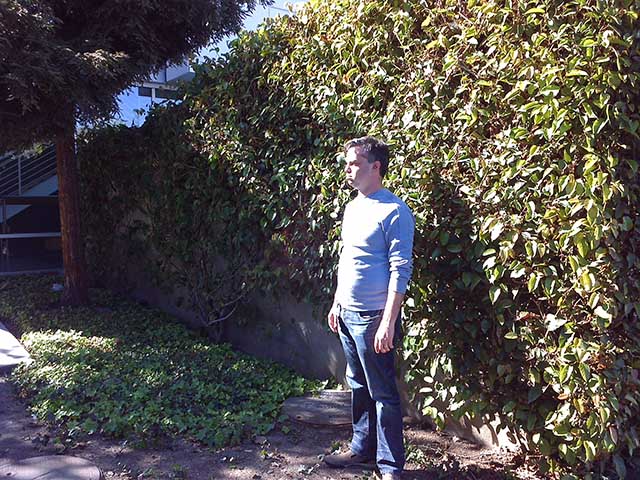}}\enskip
\subfloat[SIFT]{\label{fig:siftMatches2}\includegraphics[width=0.238\textwidth]{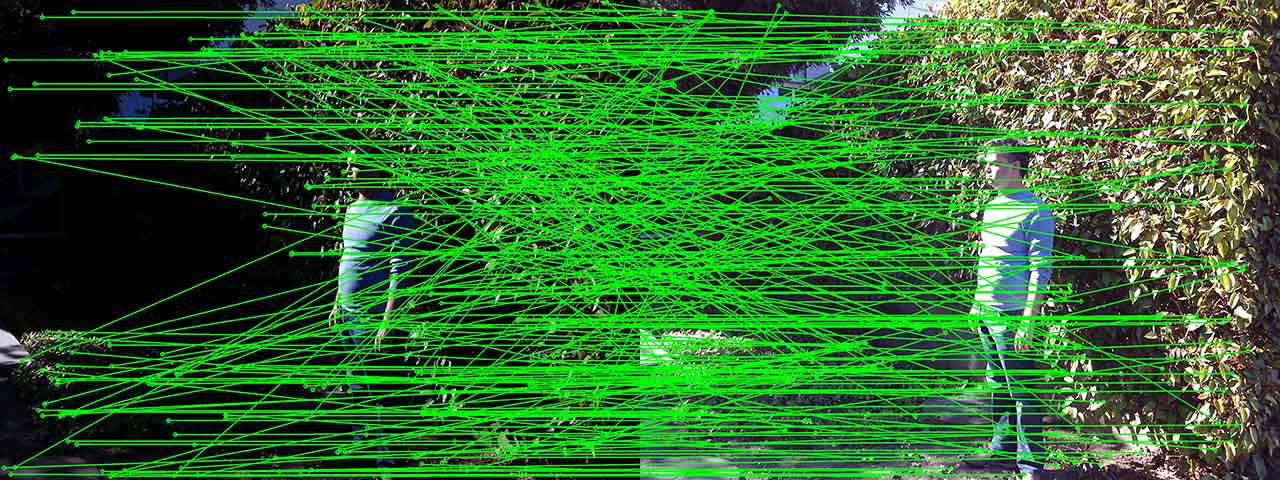}}\enskip
\subfloat[SIFT clean]{\label{fig:siftMatchesClean2}\includegraphics[width=0.238\textwidth]{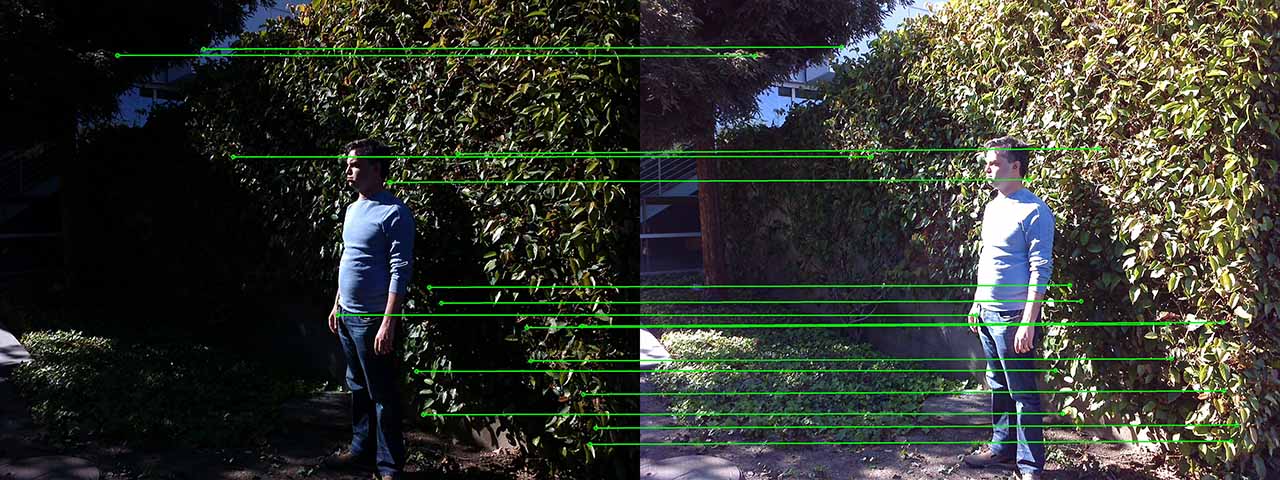}}\enskip
\subfloat[Ours]{\includegraphics[width=0.238\textwidth]{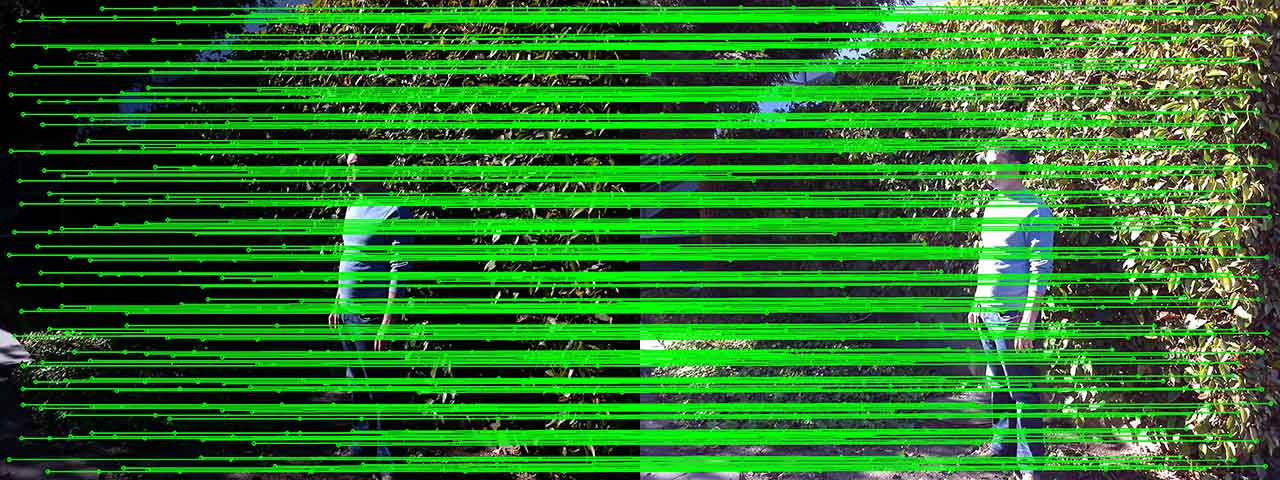}}\\
\caption{The matcher we propose performs particularly well when searching for correspondences in extremely dark areas, as is needed for large portions of the two stacks shown here. SIFT fails to find reliable correspondences; a solution could be to only retain the matches that support a homography, here indicated as ``SIFT clean''. However, if the quality of the original matches is too low, very few correspondences survive the cleaning stage, as is the case shown in (d) and (i).  Our method produces a uniformly distributed set of matches. Note that both methods were fed images that were histogram equalized.}\label{fig:siftMatches}
\end{figure*}

\subsection{Quality comparisons}\label{sec:qualityComparison}
We are particularly interested in evaluating the quality of our matcher, and the quality of the final result when compared against other non-rigid registration methods.
\vspace{2mm}

\begin{figure*}%
\centering
\newlength{\rlen}\setlength{\rlen}{0.16\textwidth}
\subfloat{\includegraphics[width=\rlen]{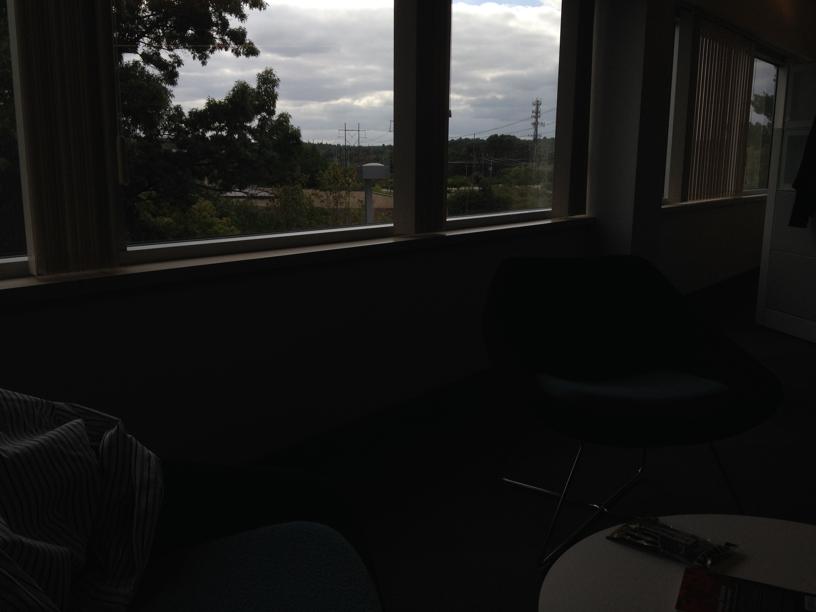}}\hfil
\subfloat{\includegraphics[width=\rlen]{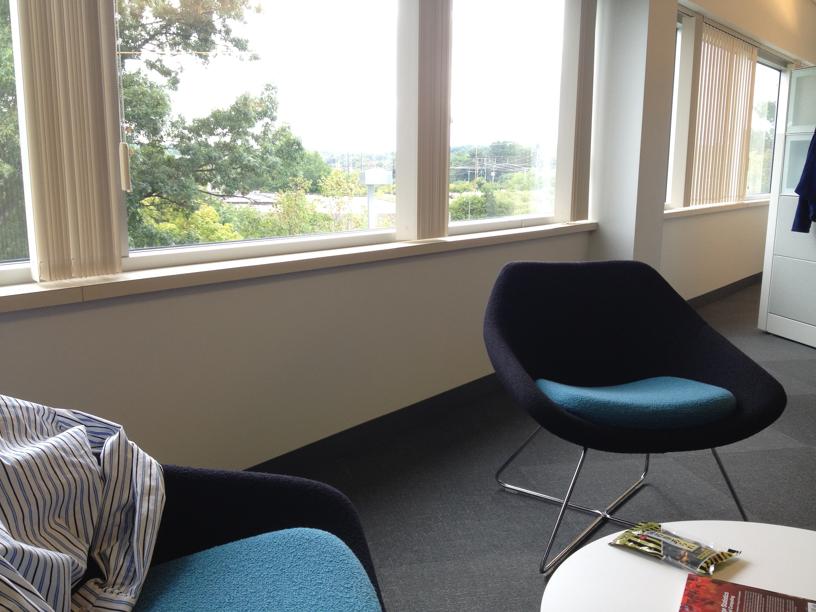}}\hfil
\subfloat{\includegraphics[width=\rlen]{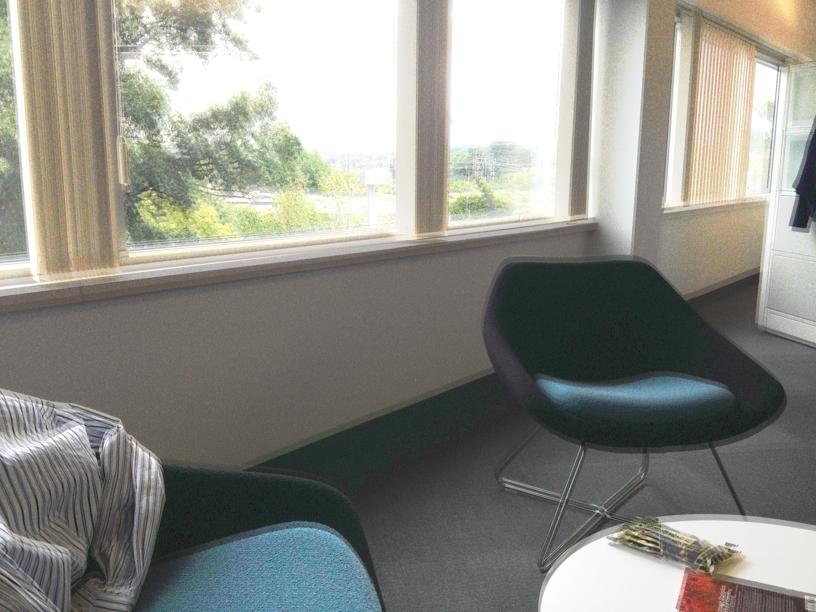}}\hfil
\subfloat{\includegraphics[width=\rlen]{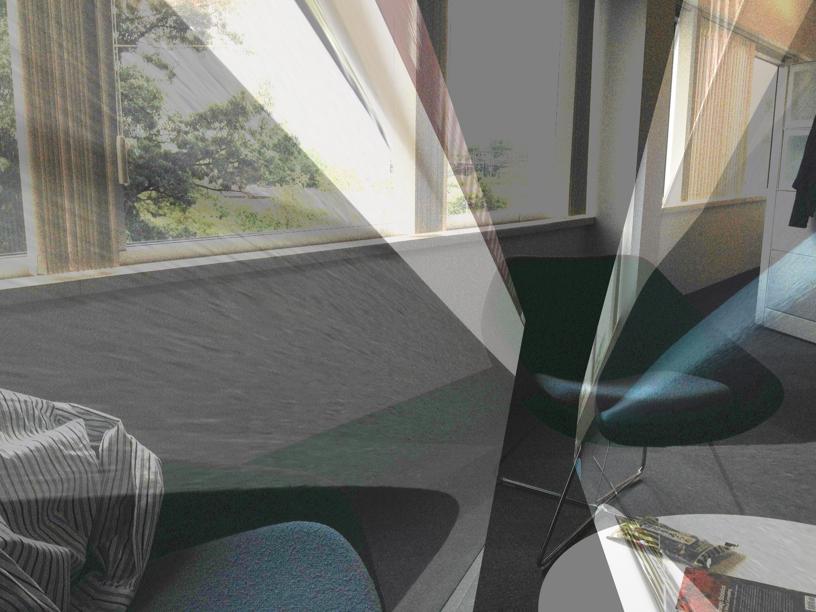}}\hfil
\subfloat{\includegraphics[width=\rlen]{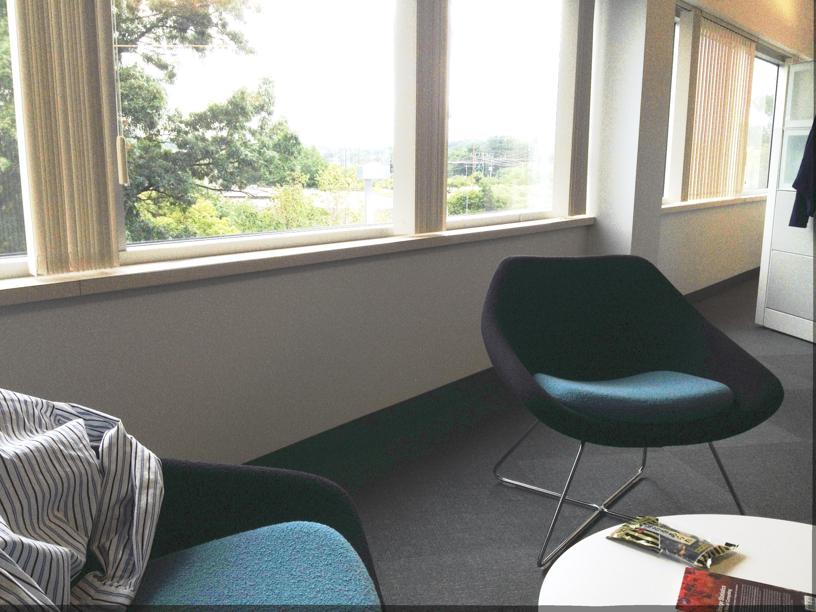}}\hfil
\subfloat{\includegraphics[width=\rlen]{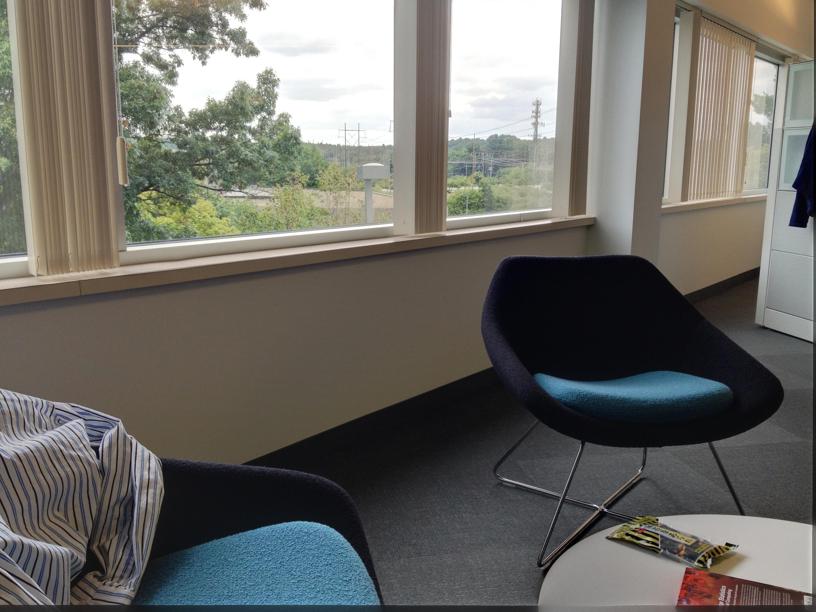}}\\[-6pt]
\subfloat{\includegraphics[width=\rlen]{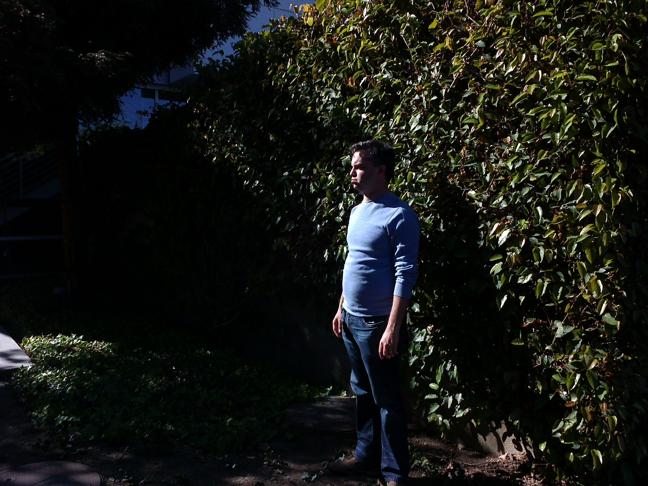}}\hfil
\subfloat{\includegraphics[width=\rlen]{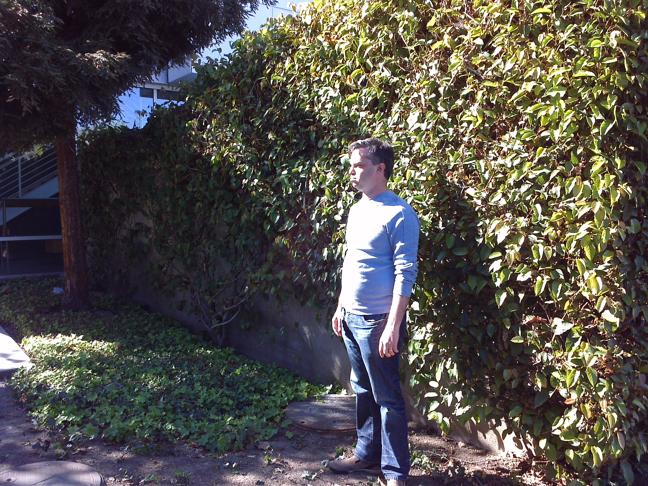}}\hfil
\subfloat{\includegraphics[width=\rlen]{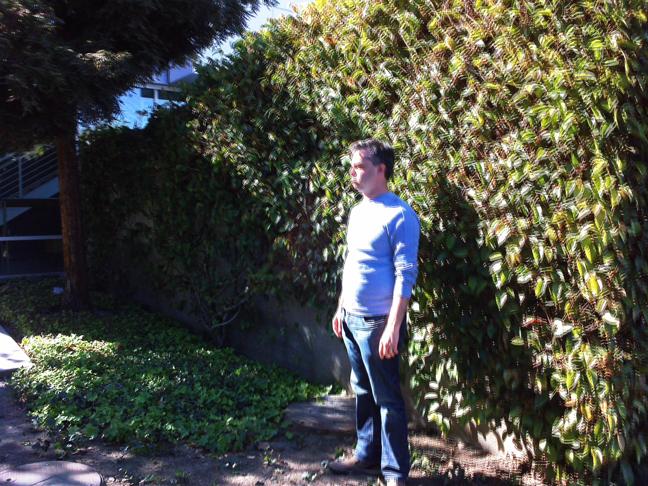}}\hfil
\subfloat{\includegraphics[width=\rlen]{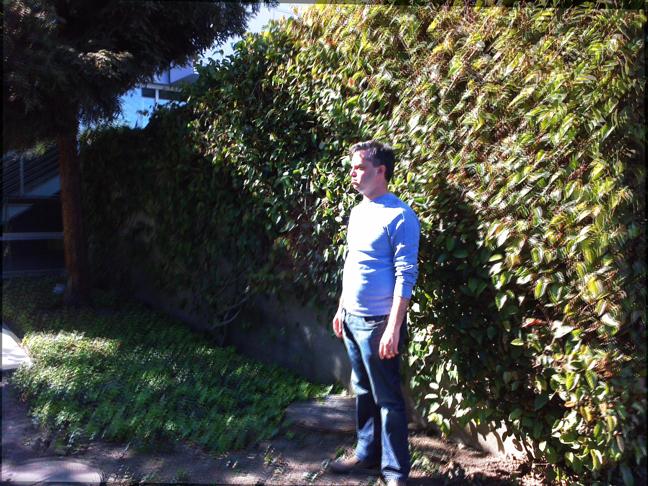}}\hfil
\subfloat{\includegraphics[width=\rlen]{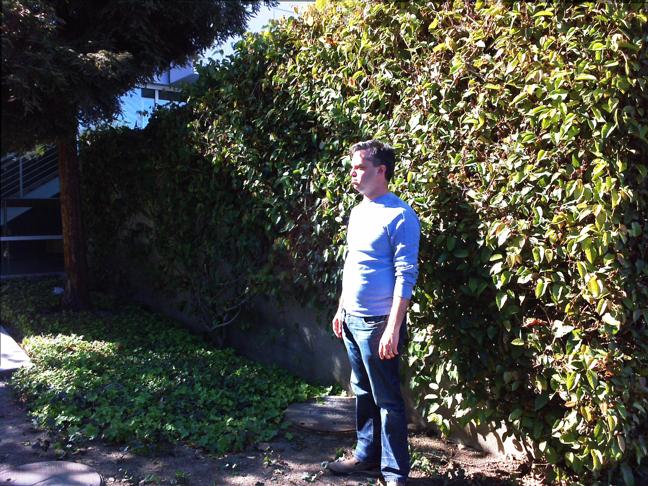}}\hfil
\subfloat{\includegraphics[width=\rlen]{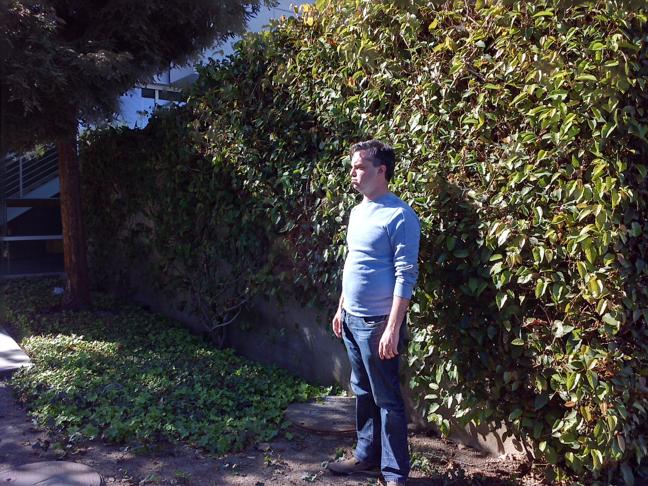}}\\[-6pt]
\subfloat{\includegraphics[width=\rlen]{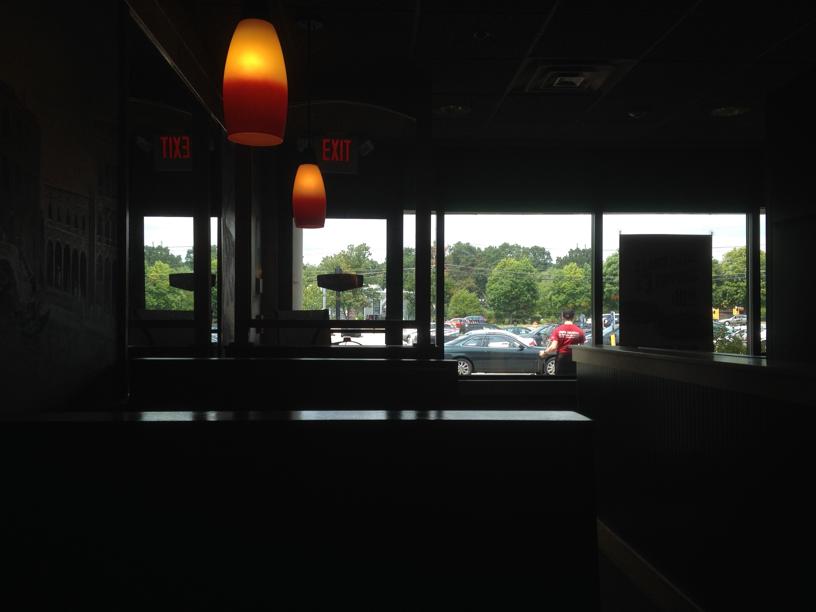}}\hfil
\subfloat{\includegraphics[width=\rlen]{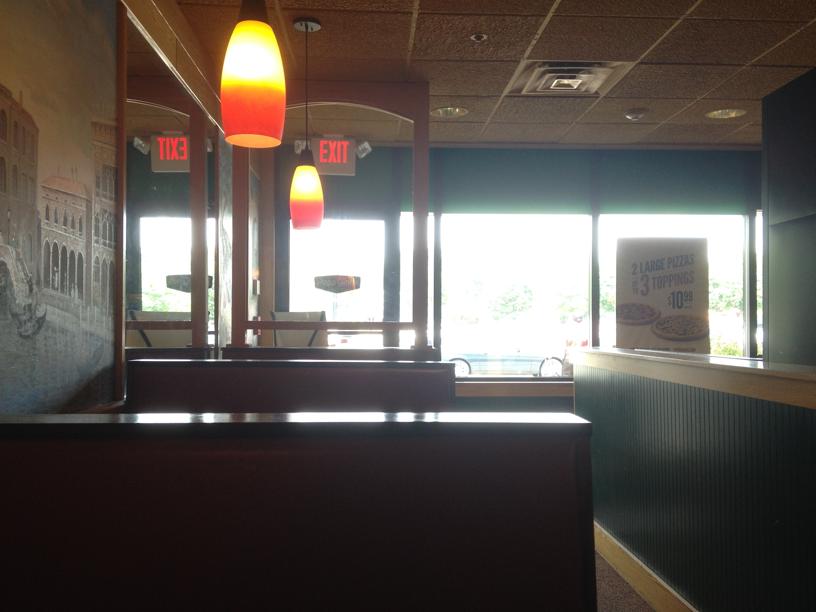}}\hfil
\subfloat{\includegraphics[width=\rlen]{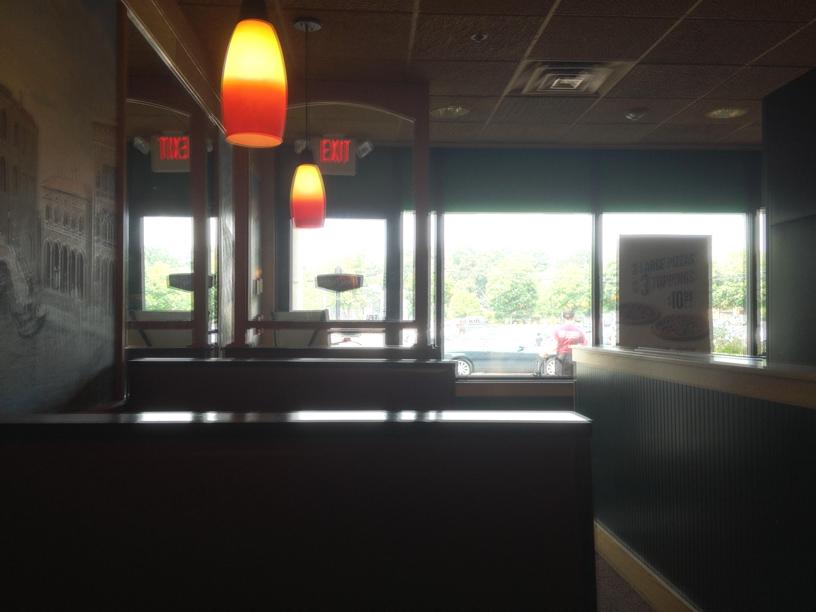}}\hfil
\subfloat{\includegraphics[width=\rlen]{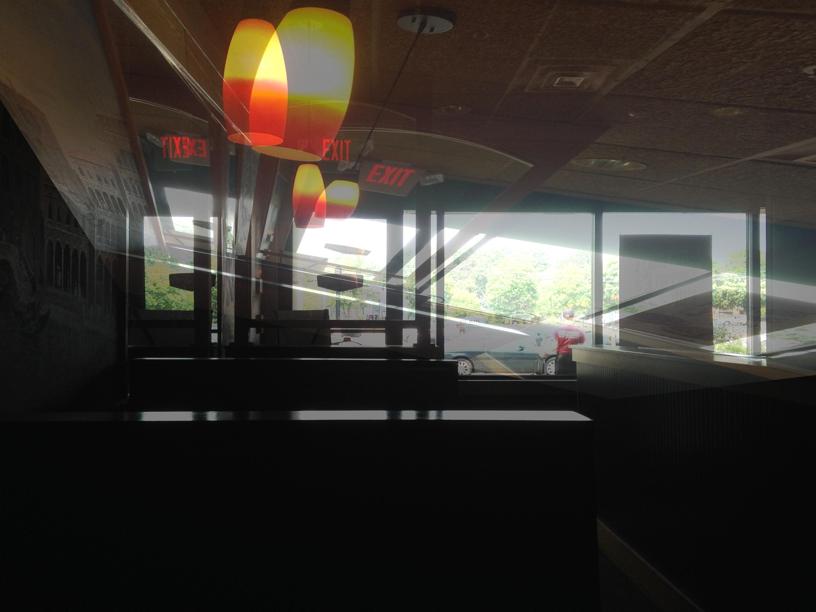}}\hfil
\subfloat{\includegraphics[width=\rlen]{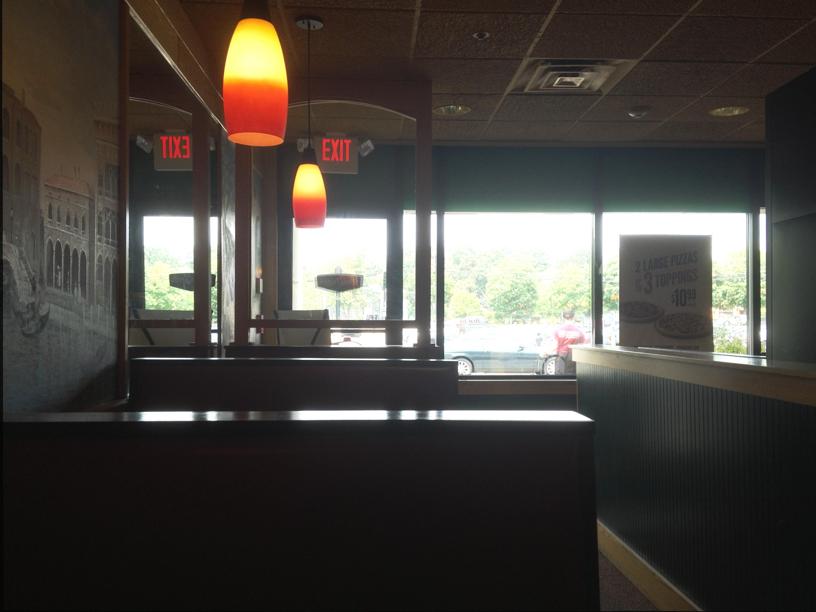}}\hfil
\subfloat{\includegraphics[width=\rlen]{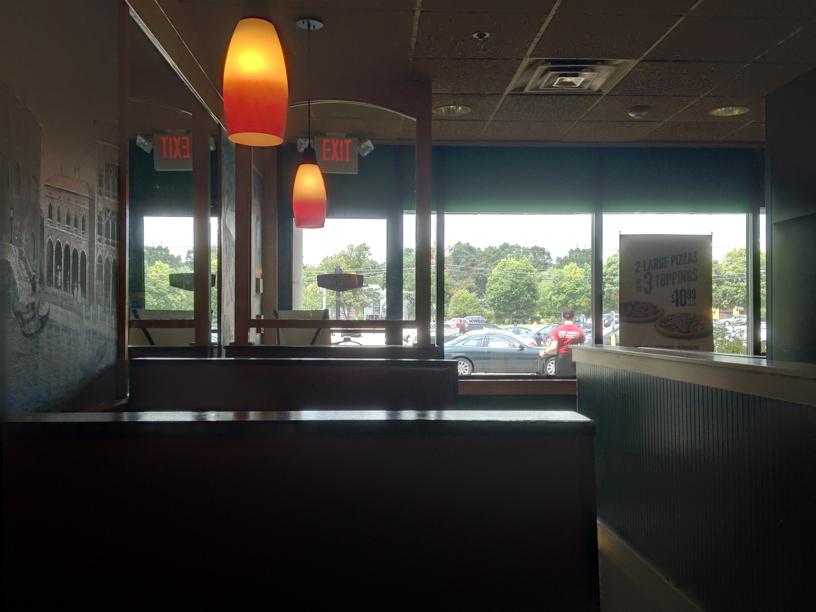}}\\[-6pt]
\setcounter{subfigure}{0}%
\subfloat[Reference]{\includegraphics[width=\rlen]{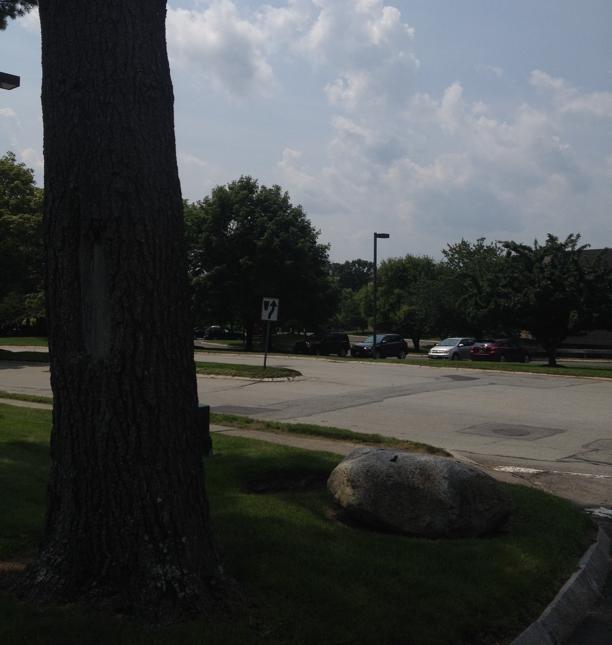}}\hfil
\subfloat[Source]{\includegraphics[width=\rlen]{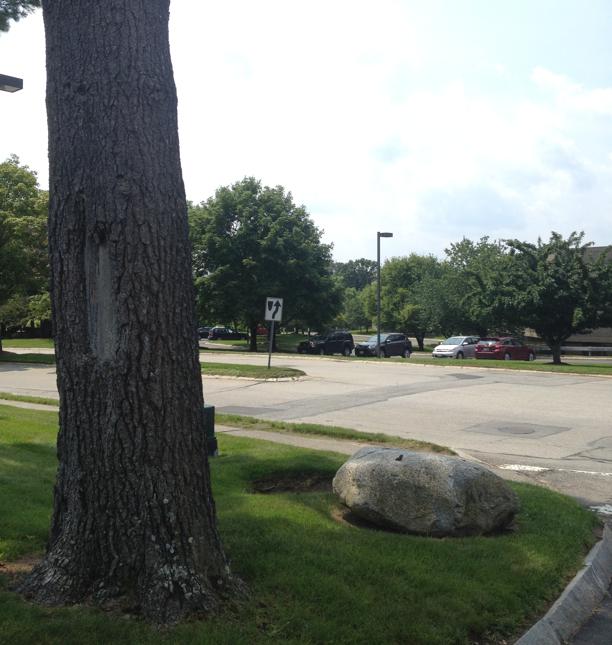}}\hfil
\subfloat[Blended Stack]{\includegraphics[width=\rlen]{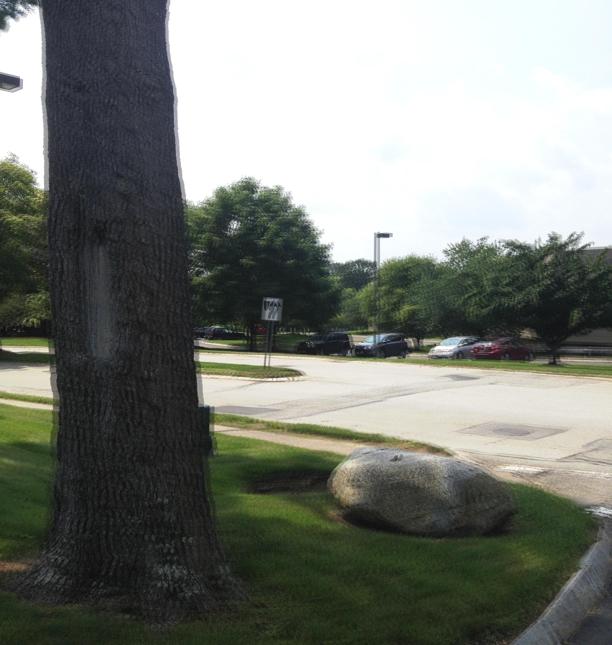}}\hfil
\subfloat[Homogr.~+ Blend]{\label{fig:hAndBlend}\includegraphics[width=\rlen]{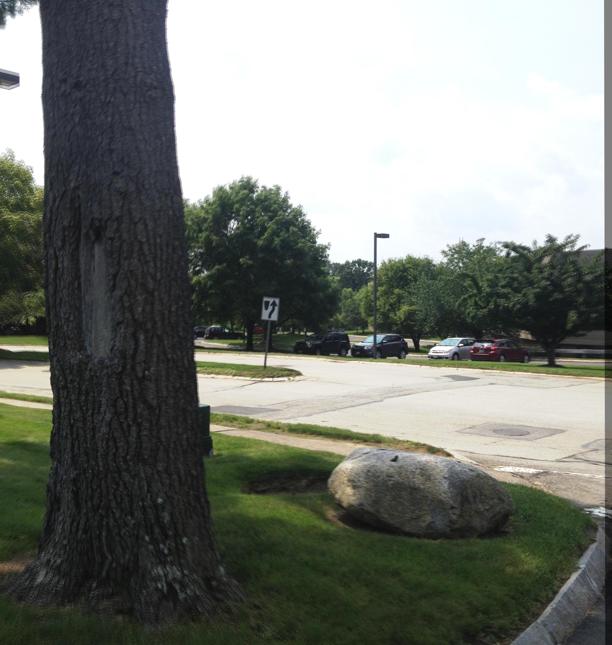}}\hfil
\subfloat[Our Align.~+  Blend]{\includegraphics[width=\rlen]{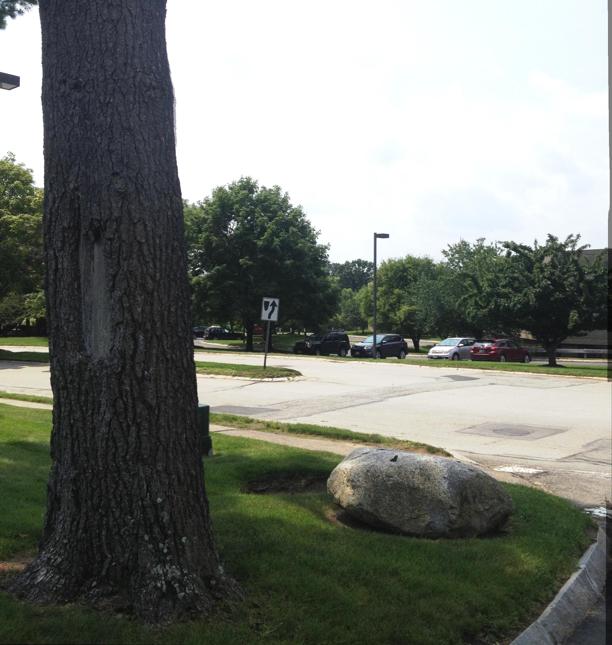}}\hfil
\subfloat[Our Final  HDR]{\includegraphics[width=\rlen]{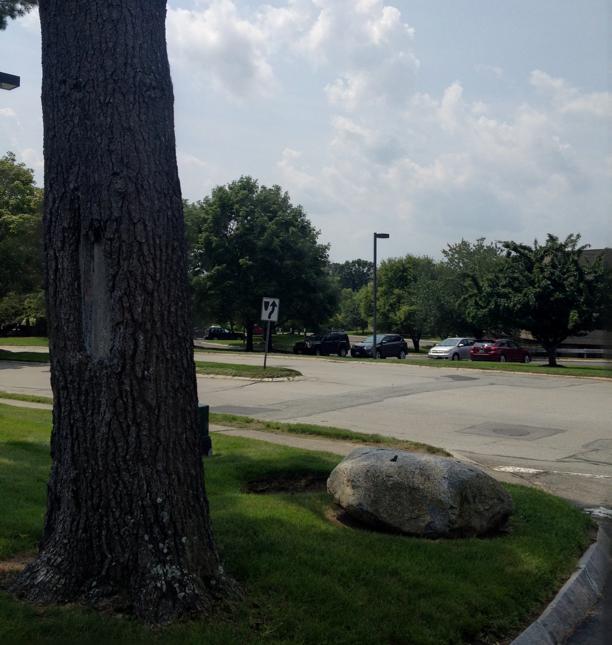}}\\
\caption{Our method works well on a variety of scenes. In (a) and (b) we show the two input images. Column (c) shows the stack directly blended together, without further alignment. In (d), we matched SIFT features and computed a single homography with RANSAC to align the images; to better show the displacements they are simply blended. Column (e) shows our alignment of the two images, again visualized using a simple blend. Finally, column (f) shows the resulting HDR image.}\label{fig:bigComparison}
\end{figure*}

\noindent\emph{The matcher}---One of our claims pertains to the robustness of our matcher in particularly low-light situations. Figure~\ref{fig:siftMatches} shows a comparison between our matcher and SIFT~\cite{Lowe1999SIFT}. In terms of robustness, SIFT is arguably the state-of-the-art method for finding correspondences between images. And indeed it can produce reliable correspondences even in the presence of large displacements, where our matcher would fail. However, when one of the two images is extremely dark, SIFT may fail dramatically, as shown in Figures~\ref{fig:siftMatches1}~ and~\ref{fig:siftMatches2}. One can filter them in a manner similar to the one we propose in Section~\ref{sec:cleaning}. Nevertheless, in extreme cases such as those shown in the figure, the correspondences may be so poor that after the filtering stage, too few are left to perform an accurate warp; the low number and quality of the correspondences shown in Figures~\ref{fig:siftMatchesClean1}~and~\ref{fig:siftMatchesClean2} for instance, are the cause of the artifacts visible in the first and second row of Figure~\ref{fig:hAndBlend}. On the contrary, our method still produces high-quality correspondences. This ability is key to the success of the registration of HDR stacks.
\vspace{2mm}

\noindent\emph{Non-rigid registration algorithms}---The context of our method is different from that of algorithms that aim at achieving a high-quality result, even in the presence of large displacements. However, we still compare on cases that are within the scope of our paper; for the comparison we pick the algorithms that can deliver the best quality~\cite{Sen2012,Hu2013}, and the fastest non-rigid registration algorithm of which we are aware~\cite{Bao2014}.

\begin{figure}%
\centering
\setcounter{subfigure}{0}
\subfloat[Reference]{\includegraphics[width=0.45\columnwidth]{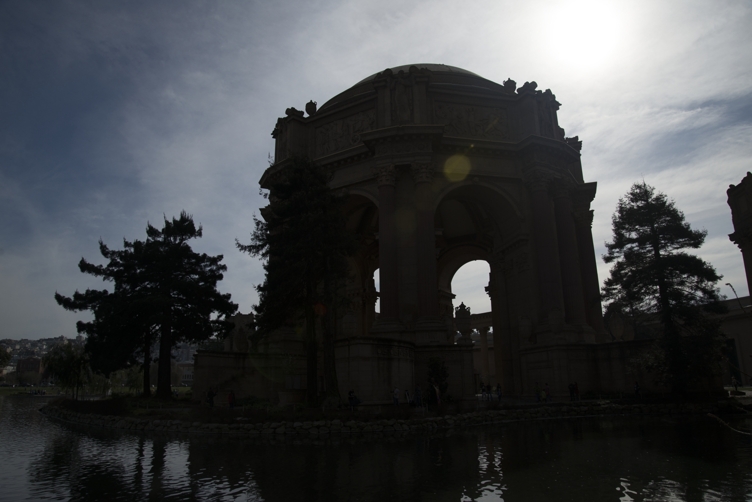}}\hspace{1mm}
\setcounter{subfigure}{3}
\subfloat[Sen~\etal~\cite{Sen2012}]{\includegraphics[width=0.45\columnwidth]{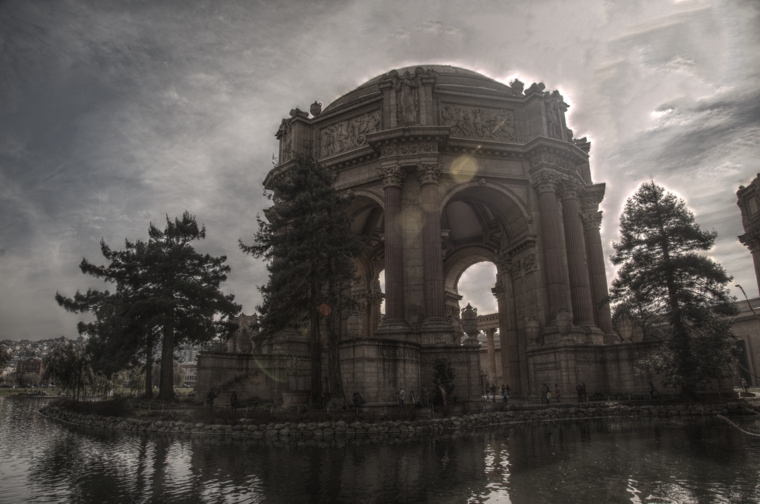}}
\vspace{-2mm}
\\
\setcounter{subfigure}{1}
\subfloat[Source 1]{\includegraphics[width=0.45\columnwidth]{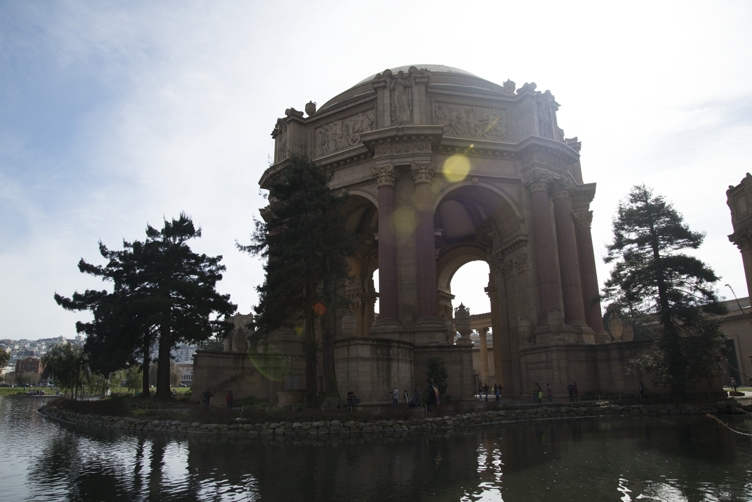}}\hspace{1mm}
\setcounter{subfigure}{4}
\subfloat[Hu~\etal~\cite{Sen2012}]{\includegraphics[width=0.45\columnwidth]{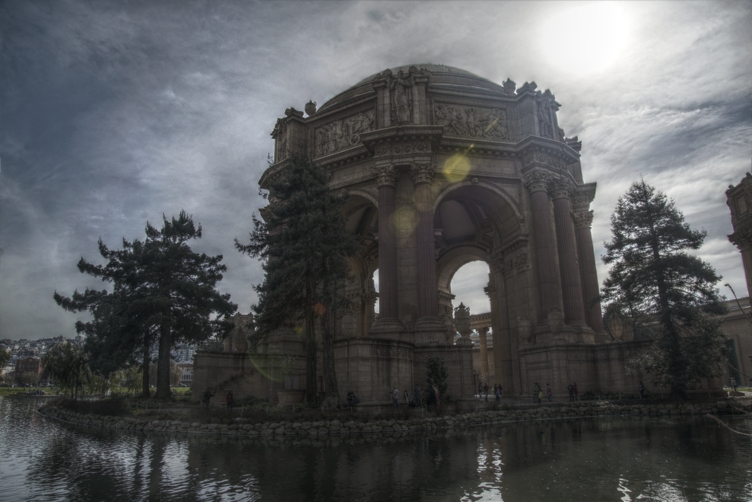}}
\\
\vspace{-2mm}
\setcounter{subfigure}{2}
\subfloat[Source 2]{\includegraphics[width=0.45\columnwidth]{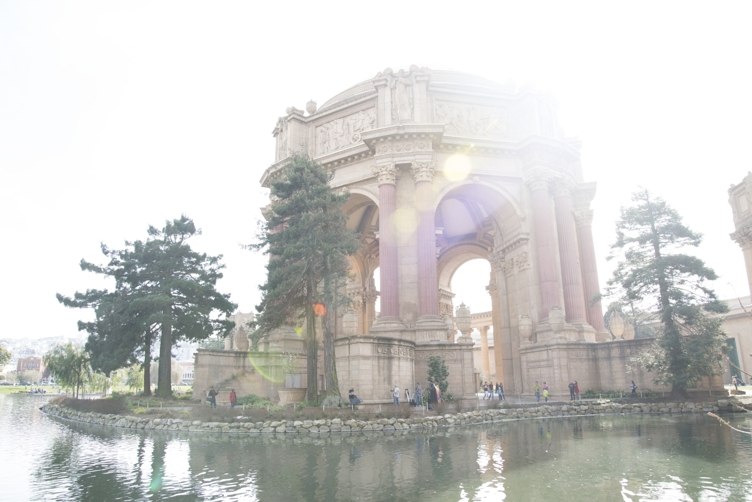}}\hspace{1mm}
\setcounter{subfigure}{5}
\subfloat[Our result]{\label{fig:ourHdr}\includegraphics[width=0.45\columnwidth]{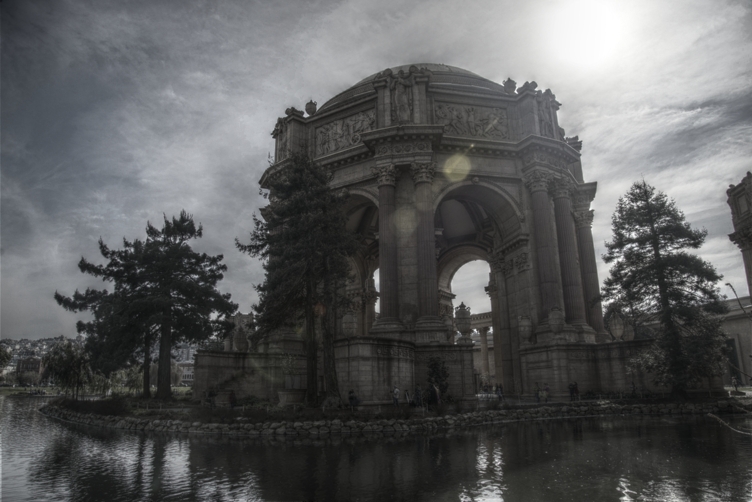}}
\caption{Comparison with the state-of-the-art methods for non-rigid registration. To produce a result that is visually comparable to the related work, we use the tonemapping operator proposed by Mantiuk~\etal~\cite{Mantiuk2006}, rather than our modified exposure fusion, see Section~\ref{sec:qualityComparison}; however, the color differences that are still visible are solely due to the tonemapper parameters.}\label{fig:fullNonRigidComparison}
\end{figure}

Figure~\ref{fig:fullNonRigidComparison} shows a comparison with competitors that deliver the highest quality. To perform it, we registered the images in the stack to the shortest exposure. Note that ``Source 2'' is $4$ stops brighter than the reference, and yet our method correctly warps it; the other two methods use the middle exposure as the reference. Also, to simplify the task of visually comparing the results of the three approaches, rather than using the modified version of exposure fusion that we described in Section~\ref{sec:ssim}, we output the warped images, create an HDR irradiance map, and use the tonemapper proposed by Mantiuk~\etal~\cite{Mantiuk2006}. The quality of the sky in our result is comparable with that of Hu~\etal, and better than that of Sen~\etal---the sun is still present and there are no halos. Note that some of the people walking under the dome are not correctly registered by our method; both the other results correctly register that region. However, as mentioned above, in this example we did not run our error-tolerant fusion, which would take care of that problem.

A method more similar in spirit to ours, is the flow algorithm recently proposed by Bao~\etal~\cite{Bao2014}. While not specifically designed for HDR registration, their algorithm is impressively fast (see Section~\ref{sec:speedComparison}). At its core, the method by Bao and colleagues uses PatchMatch to deal with large displacements~\cite{Barnes2010}.
To ameliorate the flow accuracy in occlusion and disocclusion regions, they compute the matching cost in an edge-aware fashion; at the same time they improve on speed by computing the cost only at a wisely selected subset of pixels.
Note that, despite being several times faster than the competitors, the method by Bao and colleagues ranks within the top ten positions in all of the established flow benchmark datasets.
We compare our method with theirs on cases that are within the scope of both algorithms.

Figure~\ref{fig:teaser} shows a fairly common case for an HDR stack, with both camera motion and slight scene motion (the woman is holding the volleyball). In all the comparisons with their method, we first equalize the images to compensate for illumination changes. The method by Bao and colleagues produces strong artifacts that are visible in Figure~\ref{fig:teaser}(c); on the contrary our method registers the images perfectly. Note that the original images are $5$MP images, which is possibly larger than what their method was originally designed for; please see the additional material for more comparisons, including lower resolution stacks.

\begin{figure}%
\centering
\subfloat[The stack]{\includegraphics[width=0.19\columnwidth]{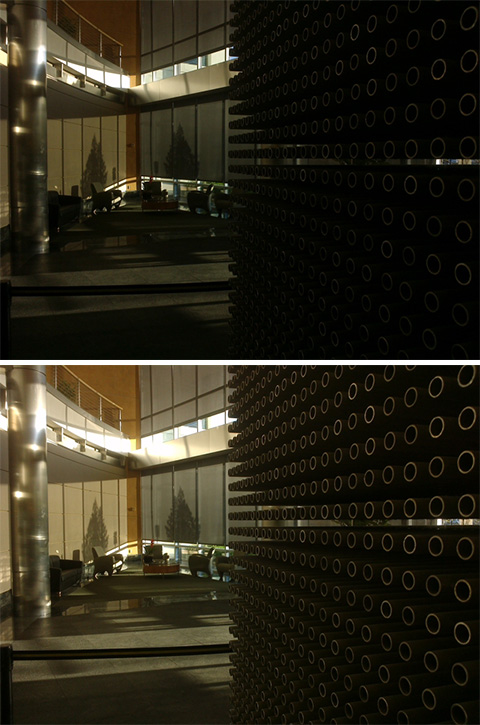}}\hspace{1mm}
\subfloat[Bao~\etal~\cite{Bao2014}]{\includegraphics[width=0.38\columnwidth]{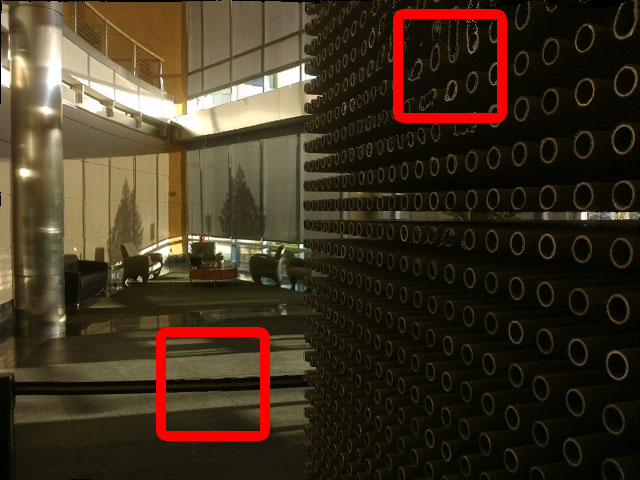}}\hspace{1mm}
\subfloat[Our result]{\includegraphics[width=0.38\columnwidth]{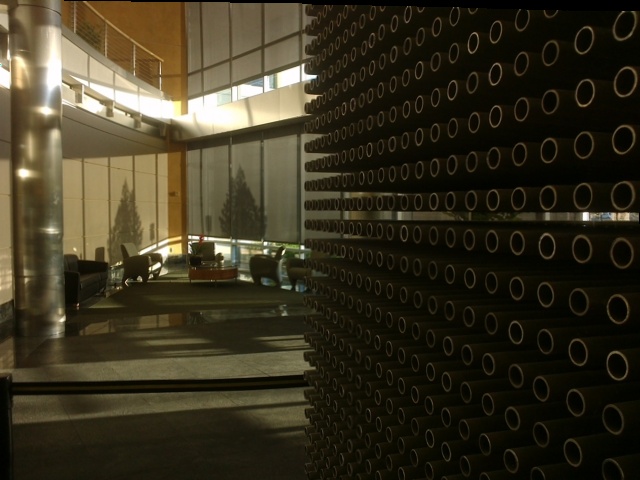}}\\
\vspace{-2mm}
\subfloat[Bao~\etal~\cite{Bao2014}]{\includegraphics[width=0.485\columnwidth]{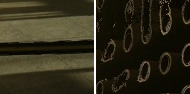}}\enskip
\subfloat[Ours]{\includegraphics[width=0.485\columnwidth]{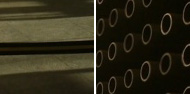}}
\caption{Comparison with the method by Bao~\etal~\cite{Bao2014}. Images (b) and (c) are the source images warped with the method by Bao~\etal and by our algorithm respectively. Notice the artifacts affecting the results by Bao~\etal.}\label{fig:bao}
\end{figure}

Figure~\ref{fig:bao} shows another comparison, this time with both algorithms running on a VGA stack. In order to perform a more fair comparison, the images were taken with a small, $1$-stop separation, and neither of them presents saturation; because of the limited dynamic range and spacing of the exposure times of this example, histogram equalization makes the source and the reference essentially identical in terms of brightness. The insets of the figure show that the method by Bao~\etal fails in preserving the local structure of the tubes.

On both examples, our algorithm produces a more accurate registration. Figure~\ref{fig:bigComparison} shows more results of our method.

\subsection{Execution time}\label{sec:speedComparison}
One of the biggest strengths of our method is its computational efficiency. We first validate this claim by comparing the runtime of our algorithm to three related works. For this experiment, we used VGA images. Two preliminary comments are in order; first, the methods by Sen~\etal and Hu~\etal are implemented in a mixture of Matlab and C++ code, which makes them intrinsically slower. However, the speedup is significant even when accounting for that. Second, the execution times shown in Table~\ref{tab:timingComparisons} for their methods are those reported by Oh~\etal~\cite{Oh2014}.

A more interesting comparison is with the method of Bao and colleagues, both because they implemented their algorithm very efficiently in CUDA, and because execution speed is one of their main focuses. Indeed, recall that, to the best of our knowledge, theirs is the fastest published method for optical flow. And yet, our method is roughly $3.5\times$ faster. This, however, is only a partial evaluation: while the execution time of our algorithm grows sublinearly, theirs grows linearly with the number of pixels, as shown in Figure~\ref{fig:bao}. On an NVIDIA GTX Titan, for a pair of 5MP images, their code runs in $1.66$s; our method registers the same images in $150$ms, which translates to a speedup $\mathbf{11\times}$.

Table~\ref{tab:timingResults} shows the cost of each step our algorithm on a desktop machine as well as a tablet, both for pairs of 5MP images. Aside from rigid registration methods, we are not aware of any published work capable of registering two $5$MP images in a time even close to a second on a desktop. Our approach can do it in less than that ($677$ms) on a commercial tablet.

\begin{figure}
\centering
\includegraphics[width=.8\columnwidth]{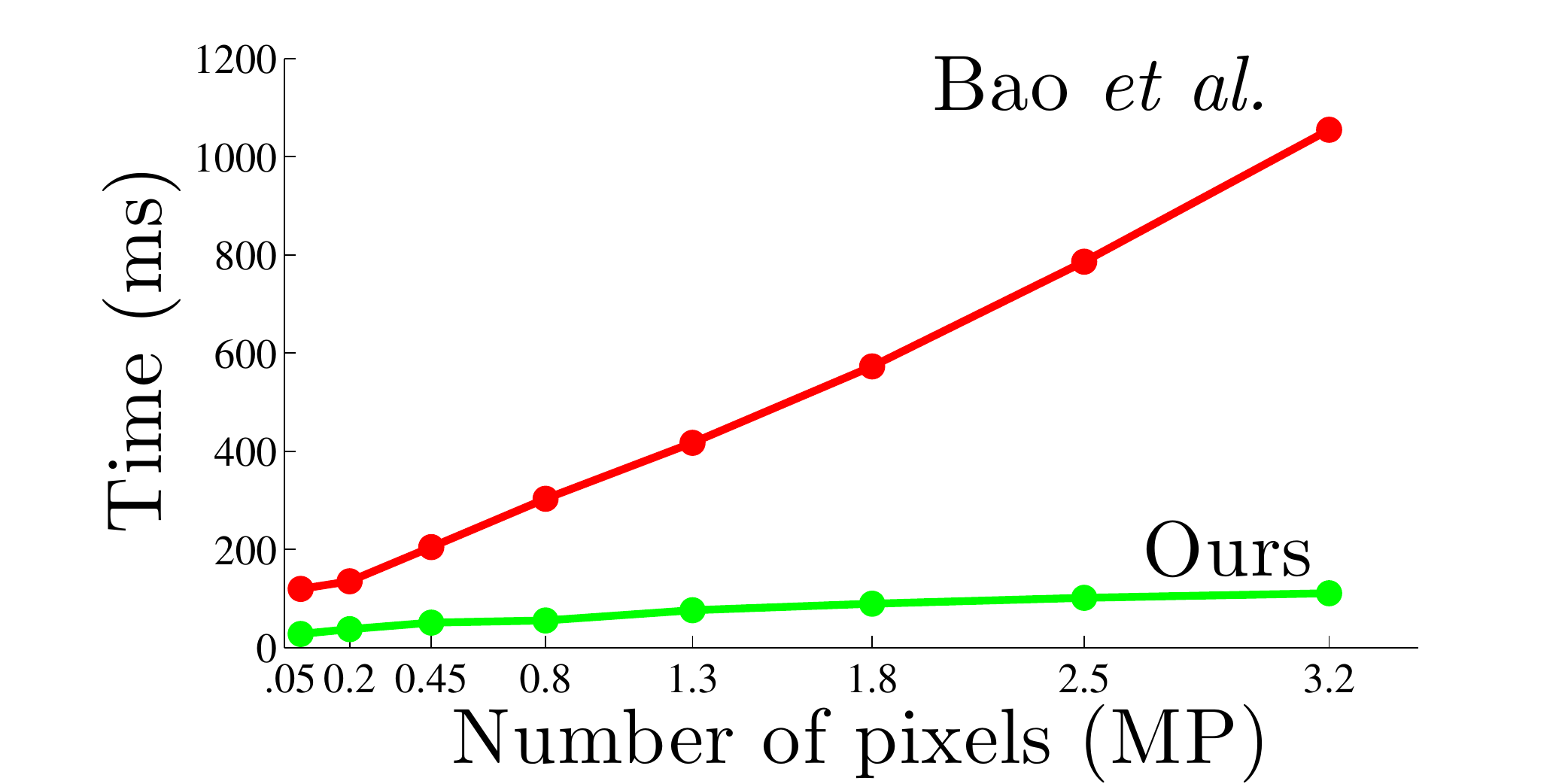}
\caption{Computational time of the algorithm by Bao~\etal~\cite{Bao2014} and ours. Note that our method grows sublinearly. The timings were captured on an NVIDIA GTX Titan graphics card.}\label{fig:baoPlot}
\end{figure}

Moreover, as shown in Figure~\ref{fig:baoPlot}, our method scales well with image size; this is a particularly attractive feature, given the rate at which the number of pixels in widely available sensors is growing.

\begin{table}
\begin{center}
\begin{tabular}{p{0.25\columnwidth}<{\raggedright\arraybackslash}>{\centering\arraybackslash}p{0.35\columnwidth}>{\centering\arraybackslash}p{0.2\columnwidth}}
Algorithm & Execution time & Speedup\\
\hline
Our algorithm & $\mathbf{49}$\textbf{ms} & ---\\
\hline
Bao~\etal~\cite{Bao2014}& $171$ms & $\approx3.5\times$\\
Sen~\etal~\cite{Sen2012} & $106^{*}$s & $>1,900^{*}\times$\\
Hu~\etal~\cite{Hu2013} & $94^{*}$s & $>2,000^{*}\times$\\
\hline
\end{tabular}
\caption{Comparison of the execution time with different state-of-the-art algorithms. The tests were run on \textbf{VGA} images. The $^*$ indicates execution times for a mixture of Matlab and C++ code.}\label{tab:timingComparisons}
\end{center}
\end{table}

\begin{table}
\begin{center}
\begin{tabular}{p{0.55\columnwidth}<{\raggedright\arraybackslash}>{\centering\arraybackslash}p{0.15\columnwidth}>{\centering\arraybackslash}p{0.15\columnwidth}}
Step of the algorithm & Tablet & Desktop\\
\hline
Matcher (Sec.~\ref{sec:matcher}) & $132$ms & $49$ms\\
Match weeding (Sec.~\ref{sec:cleaning}) & $23$ms & $20$ms\\
Sparse-to-dense flow (Sec.~\ref{sec:flowPropagation}) & $473$ms & $67$ms\\
Fusion weights (Sec.~\ref{sec:ssim}) &$49$ms&$11$ms\\
\hline
\textbf{Total time} & $\mathbf{677}$\textbf{ms} & $\mathbf{147}$\textbf{ms}
\end{tabular}
\vspace{1mm}
\caption{Computational time for each step of the algorithm when run on a pair of \textbf{5MP} images. The reference tablet is an NVIDIA Shield Tablet, which is equipped with a Tegra K1 system-on-chip. The timings on desktop were measured on an Intel I7 CPU with an NVIDIA GTX Titan graphics card.}\label{tab:timingResults}
\end{center}
\end{table}

\section{Conclusions}
In the space of registration for HDR imaging, and stack-based photography in general, it is difficult to find an acceptable trade-off between registration accuracy and computational load. 
We propose a new compromise: rather than attempting to solve the most general non-rigid registration case, we focus on the more typical case of relatively small displacement, and propose a locally non-rigid registration technique.
Specifically, we contribute a method that is $11\times$ faster than the fastest published method, while producing a more accurate registration. Our approach is also the only one that can perform non-rigid registration within the computational power of a mobile device. To achieve this result, we developed a novel, fast feature matcher that works better than the state-of-the-art when the reference image is underexposed. Our matcher comprises an original light-weight corner detector, and a matching strategy based on a modification of the RANSAC algorithm. We think that this matcher may be useful for other applications as well. Finally, we implement the complete system, from capture to HDR generation, on an NVIDIA SHIELD Tablet. This also involves a metering strategy, a flow propagation step, and a deghosting strategy to compensate for errors in the flow propagation.
vspace{-1mm}

\section*{Acknowledgments}
The authors would like to thank Colin Tracey and Huairuo Tang for their help in writing part of the early CUDA implementation.

{\small
\bibliographystyle{ieee}
\bibliography{registration}

\begin{thebibliography}{10}\itemsep=-1pt

\bibitem{Bao2014}
L.~Bao, Q.~Yang, and H.~Jin.
\newblock Fast edge-preserving {P}atch{M}atch for large displacement optical
  flow.
\newblock In {\em CVPR}, 2014.

\bibitem{Barnes2010}
C.~Barnes, E.~Shechtman, D.~B. Goldman, and A.~Finkelstein.
\newblock The generalized {PatchMatch} correspondence algorithm.
\newblock In {\em ECCV}, 2010.

\bibitem{RANSAC}
M.~A. Fischler and R.~C. Bolles.
\newblock Random sample consensus: a paradigm for model fitting with
  applications to image analysis and automated cartography.
\newblock {\em Communications of the ACM}, 1981.

\bibitem{Gallo2009}
O.~Gallo, N.~Gelfand, W.~Chen, M.~Tico, and K.~Pulli.
\newblock Artifact-free high dynamic range imaging.
\newblock In {\em ICCP}, 2009.

\bibitem{Gallo2012}
O.~Gallo, M.~Tico, R.~Manduchi, N.~Gelfand, and K.~Pulli.
\newblock Metering for exposure stacks.
\newblock {\em Eurographics}, 31:479--488, 2012.

\bibitem{Gastal2011}
E.~S. Gastal and M.~M. Oliveira.
\newblock Domain transform for edge-aware image and video processing.
\newblock In {\em ACM Transactions On Graphics}, volume~30, 2011.

\bibitem{Granados2010}
M.~Granados, B.~Ajdin, M.~Wand, C.~Theobalt, H.-P. Seidel, and H.~Lensch.
\newblock Optimal {HDR} reconstruction with linear digital cameras.
\newblock In {\em CVPR}, 2010.

\bibitem{Hasinoff2010}
S.~W. Hasinoff, F.~Durand, and W.~T. Freeman.
\newblock Noise-optimal capture for high dynamic range photography.
\newblock In {\em CVPR}, 2010.

\bibitem{Hu2013}
J.~Hu, O.~Gallo, K.~Pulli, and X.~Sun.
\newblock {HDR} deghosting: How to deal with saturation?
\newblock In {\em CVPR}, 2013.

\bibitem{Levin2004}
A.~Levin, D.~Lischinski, and Y.~Weiss.
\newblock Colorization using optimization.
\newblock In {\em ACM Transactions On Graphics}, 2004.

\bibitem{Lowe1999SIFT}
D.~G. Lowe.
\newblock Object recognition from local scale-invariant features.
\newblock In {\em ICCV}, 1999.

\bibitem{Mantiuk2006}
R.~Mantiuk, K.~Myszkowski, and H.-P. Seidel.
\newblock A perceptual framework for contrast processing of high dynamic range
  images.
\newblock {\em ACM Transactions on Applied Perception}, 2006.

\bibitem{Mertens2007}
T.~Mertens, J.~Kautz, and F.~V. Reeth.
\newblock Exposure fusion.
\newblock In {\em Pacific Graphics}, 2007.

\bibitem{Oh2014}
T.~Oh, J.~Lee, Y.~Tai, and I.~Kweon.
\newblock Robust high dynamic range imaging by rank minimization.
\newblock {\em IEEE Transactions on Pattern Analysis and Machine Intelligence},
  2014.

\bibitem{Raman2011}
S.~Raman and S.~Chaudhuri.
\newblock Reconstruction of high contrast images for dynamic scenes.
\newblock {\em The Visual Computer}, 2011.

\bibitem{Sen2012}
P.~Sen, N.~K. Kalantari, M.~Yaesoubi, S.~Darabi, D.~B. Goldman, and
  E.~Shechtman.
\newblock Robust patch-based {HDR} reconstruction of dynamic scenes.
\newblock In {\em SIGGRAPH Asia}, 2012.

\bibitem{Tzimiropoulos2010}
G.~Tzimiropoulos, V.~Argyriou, S.~Zafeiriou, and T.~Stathaki.
\newblock Robust {FFT}-based scale-invariant image registration with image
  gradients.
\newblock {\em IEEE Transactions on Pattern Analysis and Machine Intelligence},
  2010.

\bibitem{Wang2004}
Z.~Wang, A.~C. Bovik, H.~R. Sheikh, and E.~P. Simoncelli.
\newblock Image quality assessment: from error visibility to structural
  similarity.
\newblock {\em IEEE Transactions on Image Processing}, 2004.

\bibitem{Ward2003}
G.~Ward.
\newblock Fast, robust image registration for compositing high-dynamic-range
  photographs from handheld exposures.
\newblock {\em Journal of Graphics Tools}, 2003.

\bibitem{Zhang2012}
W.~Zhang and W.-K. Cham.
\newblock Reference-guided exposure fusion in dynamic scenes.
\newblock {\em Journal of Visual Communication and Image Representation}, 2012.

\end{thebibliography}
}

\end{document}